\documentclass{article}
\usepackage[numbers]{natbib}
\pdfoutput=1
\usepackage{amsmath,amsfonts,amssymb,amsthm, color}
\usepackage{algorithm,algorithmic}
\usepackage{fullpage}
\usepackage{enumerate}
\usepackage{ltxtable}
\usepackage{hyperref}
\usepackage{nicefrac}
\usepackage{multirow}
\hypersetup{
	colorlinks,%
	citecolor=black,%
	filecolor=black,%
	linkcolor=black,%
	urlcolor=black
}
\newcolumntype{L}{>{\centering\arraybackslash} m{0.05\textwidth}}
\newcolumntype{S}{>{\centering\arraybackslash} m{0.45\textwidth}}

\usepackage{fullpage}

\usepackage{amsmath,amsfonts,amssymb,amsthm}
\usepackage{graphicx}
\usepackage{color}
\usepackage{tikz}
\usetikzlibrary{arrows,shapes}
\usetikzlibrary{patterns}

\newcommand{\secref}[1]{Section~\ref{#1}}
\newcommand{\thmref}[1]{Theorem~\ref{#1}}

\newtheorem{theorem}{Theorem}
\newtheorem{lemma}{Lemma}[section]

\newtheorem{definition}{Definition}

\newcommand{\reals}{\ensuremath{\mathbb{R}}}
\newcommand{\Z}{\ensuremath{\mathbb{Z}}}

\newcommand{\argmin}[1]{\underset{#1}{\mathrm{argmin}} \:}

\newcommand{\E}[1]{{\mathbb{E}\left[{#1}\right]}}

\newcommand{\Ep}[2]{{\mathbb{E}_{#1}\left[{#2}\right]}}

\newcommand{\ip}[2]{{\left\langle{#1},{#2}\right\rangle}}

\newcommand{\norm}[1]{\left\lVert{#1}\right\rVert}

\newcommand{\z}{z}

\newcommand{\w}{\ensuremath{\mathbf{w}}}

\newcommand{\W}{\ensuremath{\mathbf{W}}}

\newcommand{\x}{\ensuremath{\mathbf{x}}}

\newcommand{\wopt}{\w^{\star}} 

\newcommand{\breg}[3]{\Delta_{#1}\left( #2\middle| #3 \right)}
\newcommand{\mc}[1]{\mathcal{#1}}
\newcommand{\mbb}[1]{\mathbb{#1}}

\renewcommand{\L}[1]{L\left(#1\right)}

\newcommand{\Wcal}{\mathcal{W}}
\renewcommand{\W}{\mathcal{W}}
\newcommand{\Ocal}{\mathcal{O}}

\renewcommand{\Z}{\mc{Z}}

\newcommand{\Es}[2]{\mbb{E}_{#1}\left[ #2 \right]}

\newcommand{\ignore}[1]{}
\title{Better Mini-Batch Algorithms\\ via Accelerated Gradient Methods}

\begin{document}
\setlength{\parskip}{2mm}
\setlength{\parindent}{0pt}

\author{\hspace{0.3in}Andrew Cotter \\ \hspace{0.3in} \texttt{cotter@ttic.edu} \\\hspace{0.3in} TTIC \\\hspace{0.3in} Chicago, IL 60637 USA \\  \hspace{2in}
 \and 
Ohad Shamir \\ \texttt{ohadsh@microsoft.com} \\ Microsoft Research\\ Cambridge, MA 02142, USA
\and 
Nathan Srebro \\ \texttt{nati@ttic.edu}  \\ TTIC \\ Chicago, IL 60637 USA\\ \hspace{2in}
\and
Karthik Sridharan \\ \texttt{karthik@ttic.edu} \\ TTIC \\ Chicago, IL 60637 USA }
\date{}

\maketitle

\begin{abstract}
  Mini-batch algorithms have been proposed as a way to speed-up
  stochastic convex optimization problems. We study how such
  algorithms can be improved using accelerated gradient methods. We
  provide a novel analysis, which shows how standard gradient methods
  may sometimes be insufficient to obtain a significant speed-up and 
  propose a novel accelerated gradient algorithm, which deals with
  this deficiency, enjoys a uniformly superior guarantee and works
  well in practice.
\end{abstract}

\section{Introduction}

We consider a stochastic convex optimization problem of the form
$$\min_{\w\in\Wcal} L(\w),$$ where $$L(\w)=\Ep{\z}{\ell(\w,\z)},$$ and
optimization is based on an empirical sample of instances
$\z_1,\ldots,\z_m$.  We focus on objectives $\ell(\w,\z)$ that are
non-negative, convex and smooth in their first argument (i.e.~have a
Lipschitz-continuous gradient).  The classical learning application is
when $\z=(\x,y)$ and $\ell(\w,(\x,y))$ is a prediction loss.  In
recent years, there has been much interest in developing efficient
first-order stochastic optimization methods for these problems, such
as stochastic mirror descent \cite{BeckTeb03,NemirovskiJuLaSh09} and
stochastic dual averaging \cite{Nesterov09,Xiao10}. These methods are
characterized by incremental updates based on subgradients $\partial
\ell(\w,\z_i)$ of individual instances, and enjoy the advantages of
being highly scalable and simple to implement.

An important limitation of these methods is that they are inherently
sequential, and so problematic to parallelize. A popular way to
speed-up these algorithms, especially in a parallel setting, is via
\emph{mini-batching}, where the incremental update is performed on an
average of the subgradients with respect to several instances at a
time, rather than a single instance (i.e.,
$\frac{1}{b}\sum_{j=1}^{b}\partial \ell(\w,\z_{i+j})$). The gradient
computations for each mini-batch can be parallelized, allowing these
methods to perform faster in a distributed framework (see for instance
\cite{ShalSiSreCo11}). Recently, \cite{DekelGilShamXia11} has shown
that a mini-batching distributed framework is capable of attaining
asymptotically optimal speed-up in general (see also \cite{AD}).

A parallel development has been the popularization of
\emph{accelerated} gradient descent methods
\cite{Nesterov83,Nesterov05,Tseng08,Lan09}. In a deterministic
optimization setting and for general smooth convex functions, these
methods enjoy a rate of $O(1/n^2)$ (where $n$ is the number of
iterations) as opposed to $O(1/n)$ using standard methods. However, in
a stochastic setting (which is the relevant one for learning
problems), the rate of both approaches have an $O(1/\sqrt{n})$
dominant term in general, so the benefit of using accelerated methods
for learning problems is not obvious.

\begin{algorithm}[t]
\caption{Stochastic Gradient Descent with Mini-Batching (SGD)}
\begin{algorithmic}
\STATE Parameters: Step size $\eta$, mini-batch size $b$.
\STATE Input: Sample $\z_1,\ldots,\z_m$
\STATE $\w_1=0$
\FOR{$i=1$ to $n=m/b$}
    \STATE Let $\ell_i(\w_i) = \frac{1}{b}\sum_{t=b(i-1)+1}^{bi}\ell(\w_i,\z_t)$
    \STATE $\w'_{i+1}:= \w_i - \eta  \nabla\ell_i(\w_i))$
    \STATE $\w_{i+1}:=P_{\Wcal}(\w'_{i+1})$
\ENDFOR
\STATE Return $\bar{\w} = \frac{1}{n}\sum_{i=1}^{n}\w_i$
\end{algorithmic}
\label{alg:md}
\end{algorithm}

\begin{algorithm}
\caption{Accelerated Gradient Method (AG)}
\begin{algorithmic}
\STATE Parameters: Step sizes $(\gamma_i,\beta_i)$, mini-batch size $b$
\STATE Input: Sample $\z_1,\ldots,\z_m$
\STATE $\w=0$
\FOR{$i=1$ to $n=m/b$}
    \STATE Let $\ell_i(\w_i) := \frac{1}{b}\sum_{t=b(i-1)+1}^{bi}\ell(\w,\z_t)$
    \STATE $\w^{\mathrm{md}}_i := \beta_i^{-1} \w_{i} + (1 - \beta_i^{-1}) \w_{i}^{\mathrm{ag}}$
    \STATE $\w'_{i+1} :=  \w^{\mathrm{md}}_{i} - \gamma_i \nabla \ell_i(\w^{\mathrm{md}}_i)$
    \STATE $\w_{i+1} :=  P_{\Wcal}(\w'_{i+1})$
    \STATE $\w^{\mathrm{ag}}_{i+1} \leftarrow \beta_i^{-1} \w_{i+1} + (1 - \beta_i^{-1}) \w_{i}^{\mathrm{ag}}$
\ENDFOR
\STATE Return $\w^{\mathrm{ag}}_{n}$
\end{algorithmic}
\label{alg:ag}
\end{algorithm}

In this paper, we study the application of accelerated methods for
mini-batch algorithms, and provide theoretical results, a novel
algorithm, and empirical experiments. The main resulting message is
that by using an appropriate accelerated method, we obtain
significantly better stochastic optimization algorithms in terms of
convergence speed. Moreover, in certain regimes acceleration is
actually \emph{necessary} in order to allow a significant speedups.
The potential benefit of acceleration to mini-batching has been
briefly noted in \cite{DMB}, but here we study this issue in much more
depth. In particular, we make the following contributions:

\begin{itemize}
\item We develop novel convergence bounds for the standard
  gradient method, which refines the result of
  \cite{DekelGilShamXia11,DMB} by being dependent on
  $L(\wopt) = \inf_{\w \in \Wcal} L(\w)$, the expected loss of the best predictor in our class.
  For example, we show that in the regime where the desired
  suboptimality is comparable or larger than $L(\wopt)$, including in
  the separable case $L(\wopt)=0$, mini-batching does not lead to
  significant speed-ups with standard gradient methods.
\item We develop a novel variant of the stochastic accelerated
  gradient method \cite{Lan09}, which is optimized for a mini-batch
  framework and implicitly adaptive to $L(\wopt)$.
\item We provide an analysis of our accelerated algorithm,
  refining the analysis of \cite{Lan09} by being dependent on
  $L(\wopt)$, and show how it always allows for significant speed-ups
  via mini-batching, in contrast to standard gradient methods.
  Moreover, its performance is uniformly superior, at
  least in terms of theoretical upper bounds.
\item We provide an empirical study, validating our theoretical
  observations and the efficacy of our new method.
\end{itemize}

\section{Preliminaries}\label{sec:prelim}
We consider stochastic convex optimization problems over some convex
domain $\Wcal$.  Here, we take $\Wcal$ to be a convex subset of a
Euclidean space, and use $\norm{\w}$ to denote the standard Euclidean
norm.  In the Appendix, we state and prove the result in a more
general setting, where $\Wcal$ is a convex subset of a Banach space,
and $\norm{\w}$ can be an arbitrary norm.(subset of Euclidean case,
see Appendix for the more general Banach space case), using an i.i.d.
sample $z_1,\ldots,z_m \in \Z$ drawn from some fixed distribution.

Throughout this paper we assume that the instantaneous loss $\ell :
\Wcal \times \Z \mapsto \reals$ is convex in its first argument and
non-negative. We further assume that the loss is $H$-smooth in its
first argument for each $\z \in \Z$. That is for every $\z \in \Z$ and
$\w , \w' \in \Wcal$,
$$
\norm{\nabla \ell(\w,z) - \nabla \ell(\w',z)} \le H \norm{\w - \w'}
$$
(for more general Banach space case, the norm on the left hand side is the dual norm). Let us denote
$$
L(\w) := \Ep{\z}{\ell(\w,\z)}
$$
We wish to minimize $L(\w)$ over convex domain $\Wcal$.  We will
provide guarantees on $L(\w)$ relative to $L(\wopt)$ at some $\wopt
\in \Wcal$, where the guarantees also depend on $\norm{\wopt}$.  We could choose
$\wopt := \arg\min_{\w \in \W} L(\w)$, though our results hold for any
$\wopt\in\Wcal$, and in some cases we might choose to compete with a
low-norm $\wopt$ that is not optimal in $\Wcal$.

The behavior of the accelerated gradient method also depends on the
radius of $\Wcal$, defined as:
$$
D := \sup_{\w \in \Wcal} \norm{\w}
$$

We discuss two stochastic optimization approaches to deal with this
problem: stochastic gradient descent (SGD), and accelerated gradient
methods (AG). In a mini-batch setting, both approaches iteratively
average sub-gradients with respect to several instances, and use this
average to update the predictor. However, the update is done in
different ways. In the Appendix, we also provide the form of the
update in the more general mirror descent setting, where $\norm{\w}$
is an arbitrary norm.

The stochastic gradient descent algorithm is summarized as Algorithm
\ref{alg:md}. In the pseudocode, $P_{\Wcal}$ refers to the projection
on to the ball $\Wcal$ (under the Euclidean distance).  The
accelerated gradient method (e.g., \cite{Lan09}) is summarized as
Algorithm \ref{alg:ag}.

In terms of existing results, for the SGD algorithm we have
\cite[Section 5.1]{DMB}
\[
\E{L(\bar{\w})} - L(\wopt) \le \Ocal\left(\sqrt{\frac{1}{m}}+\frac{b}{m}\right),
\]
whereas for an accelerated gradient algorithm, we have \cite{Lan09}
\[
\E{L(\w^{\mathrm{ag}}_n)} - L(\wopt) \le \Ocal\left(\sqrt{\frac{1}{m}}+\frac{b^2}{m^2}\right),
\]
where in both cases the dependence on $D, H$ and $\norm{\wopt}$ is
suppressed.  The above bounds suggest that, as long as
$b=o(\sqrt{m})$, both methods allow us to use a large mini-batch size
$b$ without significantly degrading the performance of either method.
This allows the number of iterations $n=m/b$ to be smaller,
potentially resulting in faster convergence speed. However, these
bounds do not show that accelerated methods have a significant
advantage over the SGD algorithm, at least when $b=o(\sqrt{m})$, since
both have the same first-order term $1/\sqrt{m}$. To understand the
differences between these two methods better, we will need a more
refined analysis, to which we now turn.

\section{Convergence Guarantees}\label{sec:convergence}

The following theorems provide a refined convergence guarantee for the
SGD algorithm and the AG algorithm, which improves on the analysis of
\cite{DekelGilShamXia11,DMB,Lan09} by being explicitly dependent on
$L(\wopt)$, the expected loss of the best predictor $\wopt$ in $\Wcal$.

\begin{theorem}\label{thm:sgd}
  For any $\wopt\in\Wcal$, using Stochastic Gradient Descent with a
  step size of $\eta = \min\left\{\frac{1}{2H}, \tfrac{\sqrt{\frac{b
          \norm{\w^*}^2 }{L(\wopt) H n}}}{ 1 + \sqrt{\frac{H
          \norm{\w^*}^2}{L(\wopt) b n}}} \right\}$, we have:
\begin{align*}
\E{L(\bar{\w})} - L(\wopt) \le \sqrt{\frac{64 H  \norm{\wopt}^2 L(\wopt)}{b n}} +  \frac{4 L(\wopt) + 4 H \norm{\wopt}^2}{n} + \frac{8 H \norm{\wopt}^2}{b n}
\end{align*}
\end{theorem}

Note that the radius $D$ does not appear in the above bound, which
depends only on $\norm{\wopt}$.  This means that $\Wcal$ could be
unbounded, perhaps even the entire space, and a projection step for
SGD is not really crucial.  The step size, of course, still depends on
$\norm{\wopt}$.

\begin{theorem}\label{thm:ag}
For any $\wopt\in\Wcal$, using Accelerated Gradient with step size
parameters $\beta_i = \frac{i+1}{2}$, $\gamma_i = \gamma i^p$ where
\begin{align}\label{eq:optgamma}
\gamma = \min \left\{\tfrac{1}{4 H},\ \sqrt{\tfrac{b \norm{\w^*}^2}{348 H L(\wopt) (n-1)^{2p+1}}},\  \left(\tfrac{b}{1044 H (n - 1)^{2p}}\right)^{\frac{p+1}{2p+1}} \left(\tfrac{\norm{\w^*}^2 }{ 4 H \norm{\w^*}^2  + \sqrt{4 H \norm{\w^*}^2 L(\wopt)}} \right)^{\frac{p}{2p+1}}\right\}
\end{align}
and
\begin{align}\label{eq:optp}
p = \min\left\{\max\left\{\frac{\log(b)}{2 \log(n-1)} , \frac{\log \log(n)}{2\left(\log(b(n-1)) - \log \log(n)\right)}\right\} ,1\right\}~~,
\end{align}
as long as $n \ge 783$, we have:
\begin{align*}
\E{L(\w^{\mathrm{ag}}_n)} - L(\wopt) & \le 117 \sqrt{\frac{H \norm{\wopt}^2 L(\wopt)}{b n}} + \frac{367  H  \norm{\wopt}^{4/3}  D^{\frac{2}{3}} }{\sqrt{b} n}   +  \frac{546 H D^2 \sqrt{\log(n)}}{b n }  + \frac{5 H \norm{\wopt}^2 }{n^{2}} \\
&  \le 117 \sqrt{\frac{H D^2 L(\wopt)}{b n}} + \frac{367  H  D^2 }{\sqrt{b} n}   +  \frac{546 H D^2 \sqrt{\log(n)}}{b n}  + \frac{5 H D^2 }{ n^{2}} 
\end{align*}
\end{theorem}
Unlike for SGD, notice that the bound for the AG method above does
depend on $D$, and a projection step is necessary for our analysis.
However it is worth noting that $D$ only appears in terms of order at
least $1/n$, and appears only mildly in the $1/(\sqrt{b}n)$ term,
suggesting some robustness to the radius $D$.

We emphasize that \thmref{thm:ag} gives more than a theoretical bound:
it actually specifies a novel accelerated gradient strategy, where the
step size $\gamma_i$ scales {\em polynomially} in $i$, in a way
dependent on the minibatch size $b$ and $L(\wopt)$.  While $L(\wopt)$
may not be known in advance, it does have the practical implication
that choosing $\gamma_i\propto i^p$ for some $p<1$, as opposed to just
choosing $\gamma_i \propto i$ as in \cite{Lan09}), might yield
superior results. 

We now provide a proof sketch of Theorems \ref{thm:sgd} and
\ref{thm:ag}.  A more general statement of the Theorems as well as a
complete proof can be found in the Appendix.

The key observation used for analyzing the dependence on $L(\wopt)$ is
that for any non-negative $H$-smooth convex function
$f : \W \mapsto \reals$, we have \cite{SreSriTew10}:
\begin{align}\label{eq:self}
\norm{\nabla f(\w)} \le \sqrt{4 H f(\w)}
\end{align}
This self-bounding property tells us that the norm of the gradient is
small at a point if the loss is itself small at that point. This
self-bounding property has been used in \cite{Shalev07} in the online
setting and in \cite{SreSriTew10} in the stochastic setting to get
better (faster) rates of convergence for non-negative smooth losses.
The implication of this observation are that for any $\w \in \W$,
$\norm{\nabla L(\w)} \le \sqrt{4 H L(\w)}$ and $\forall \z \in \Z,
\norm{\ell(\w, \z)} \le \sqrt{4 H \ell(\w,\z)}$.

\begin{proof}[{\bf Proof sketch for Theorem \ref{thm:sgd}}]
The proof for the stochastic gradient descent bound is mainly based on the proof techniques in \cite{Lan09} and its extension to the mini-batch case in \cite{DekelGilShamXia11}. Following the line of analysis in \cite{Lan09}, one can show that
$$
\E{ \tfrac{1}{n} \sum_{i=1}^{n} L(\w_{i})} - L(\wopt) \le \tfrac{\eta}{n-1} \sum_{i=1}^{n-1} \E{\norm{\nabla L(\w_i) - \nabla \ell_i(\w_i)}^2} + \tfrac{D^2}{2 \eta (n-1)}
$$
In the case of \cite{Lan09}, $\E{\norm{\nabla L(\w_i) - \nabla \ell_i(\w_i)}}$ is bounded by the variance, and that leads to the final bound provided in \cite{Lan09} (by setting $\eta$ appropriately). As noticed in \cite{DekelGilShamXia11}, in the minibatch setting we have
$\nabla \ell_i(\w_i) = \frac{1}{b} \sum_{t=b(i-1)+1}^{bi} \ell(\w_i,\z_t)$
and so one can further show that
\begin{align}\label{eq:DGSX}
\E{ \tfrac{1}{n} \sum_{i=1}^{n} L(\w_{i})} - L(\wopt) \le \tfrac{\eta}{b^2(n-1)} \sum_{i=1}^{n-1} \sum_{\underset{(i-1)b+1}{t=}}^{ib} \mathbb{E}\norm{\nabla L(\w_i) - \nabla \ell(\w_i,\z_t)}^2 + \tfrac{D^2}{2 \eta (n-1)}
\end{align}
In \cite{DekelGilShamXia11}, each of $\norm{\nabla L(\w_i) - \nabla \ell(\w_i,\z_t)}$ is bounded by $\sigma_0$ and so setting $\eta$, the mini-batch bound provided there is obtained. In our analysis we further use the self-bounding property to \eqref{eq:DGSX} and get that
$$
\E{ \tfrac{1}{n} \sum_{i=1}^{n} L(\w_{i})} - L(\wopt) \le \tfrac{16 H \eta}{b(n-1)} \sum_{i=1}^{n-1}  \E{L(\w_i)}  + \tfrac{D^2}{2 \eta (n-1)}
$$
rearranging and setting $\eta$ appropriately gives the final bound.
\end{proof}

\begin{proof}[{\bf Proof sketch for Theorem \ref{thm:ag}}]
The proof of the accelerated method starts in a similar way as in \cite{Lan09}. For the $\gamma_i$'a and $\beta_i$'s mentioned in the theorem, following similar lines of analysis as in \cite{Lan09} we get the preliminary bound
$$
\E{L(\w^{\mathrm{ag}}_n)} - L(\wopt) \le \frac{2 \gamma}{(n-1)^{p+1}} \sum_{i=1}^{n-1} i^{2p}\ \E{\norm{\nabla L(\w^{\mathrm{md}}_i) - \nabla \ell_i(\w^{\mathrm{md}}_i)}^2} + \frac{D^2}{\gamma (n-1)^{p+1}}
$$
In \cite{Lan09} the step size $\gamma_i = \gamma (i+1)/2$ and $\beta_i = (i+1)/2$ which effectively amounts to $p = 1$ and further similar to the stochastic gradient descent analysis. Furthermore, each $\E{\norm{\nabla L(\w^{\mathrm{md}}_i) - \nabla \ell_i(\w^{\mathrm{md}}_i)}^2} $ is assumed to be bounded by some constant, and thus leads to the final bound provided in \cite{Lan09} by setting $\gamma$ appropriately. On the other hand, we first notice that due to the mini-batch setting, just like in the proof of stochastic gradient descent,
\begin{align*}
\E{L(\w^{\mathrm{ag}}_n)} - L(\wopt) & \le \tfrac{2 \gamma}{b^2 (n-1)^{p+1}} \sum_{i=1}^{n-1} i^{2p} \sum_{\underset{b(i-1)+1}{t=}}^{ib} \E{\norm{\nabla L(\w^{\mathrm{md}}_i) - \nabla \ell(\w^{\mathrm{md}}_i,z_t)}^2} + \tfrac{D^2}{\gamma (n-1)^{p+1}}
\end{align*}
Using smoothness, the self bounding property some manipulations, we can further get the bound
\begin{align*}
\E{L(\w^{\mathrm{ag}}_n)} - L(\wopt) & \le \tfrac{64 H \gamma}{b(n-1)^{1-p}} \sum_{i=1}^{n-1} \left(\E{L(\w^{\mathrm{ag}}_i)} - L(\wopt) \right) + \tfrac{64 H \gamma L(\wopt) (n-1)^p}{b} \\
& ~~~~~~~~~~ + \tfrac{D^2}{\gamma (n-1)^{p+1}} + \tfrac{32 H D^2}{b(n-1)}
\end{align*}
Notice that the above recursively bounds $\E{L(\w^{\mathrm{ag}}_n)} - L(\wopt)$ in terms of $\sum_{i=1}^{n-1} \left(\E{L(\w^{\mathrm{ag}}_i)} - L(\wopt) \right)$. While unrolling the recursion all the way down to $2$ does not help, we notice that for any $\w \in \Wcal$, $L(\w)- L(\wopt) \le 12 HD^2 + 3 L(\wopt)$. Hence we unroll the recursion to $M$ steps and use this inequality for the remaining sum. Optimizing over number of steps up to which we unroll and also optimizing over the choice of $\gamma$, we get the bound,
\begin{align*}
\E{L(\w^{\mathrm{ag}}_n)} - L(\wopt) & \le \sqrt{\tfrac{1648 H D^2 L(\wopt)}{b (n-1)}} + \tfrac{348 (6 HD^2 + 2 L(\wopt))}{b(n-1)} (b (n-1))^{\frac{p}{p+1}} + \tfrac{32 H D^2}{b(n-1)} \\
& ~~~~~ + \tfrac{4 H D^2}{(n-1)^{p+1}} + \tfrac{36 H D^2}{b(n-1)} \tfrac{\log(n)}{(b (n-1))^{\frac{p}{2p + 1}}}
\end{align*}
Using the $p$ as given in the theorem statement, and few simple manipulations, gives the final bound.
\end{proof}

\section{Optimizing with Mini-Batches}\label{sec:minibatches}

To compare our two theorems and understand their implications, it
will be convenient to treat $H$ and $D$ as constants, and focus on the
more interesting parameters of sample size $m$, minibatch size $b$,
and optimal expected loss $L(\wopt)$. Also, we will ignore the
logarithmic factor in \thmref{thm:ag}, since we will mostly be
interested in significant (i.e. polynomial) differences between the
two algorithms, and it is quite possible that this logarithmic factor is merely an artifact of our analysis. Using $m=nb$, we
get that the bound for the SGD algorithm is
\begin{equation}\label{eq:sgd}
 \E{L(\bar{\w})}  - L(\wopt) ~\le~ \tilde{\Ocal}\left(\sqrt{\frac{L(\wopt)}{bn }}+\frac{1}{n}\right) ~=~ \tilde{\Ocal}\left(\sqrt{\frac{L(\wopt)}{m }}+\frac{b}{m}\right),
\end{equation}
and the bound for the accelerated gradient method we propose is
\begin{equation}\label{eq:ag}
\E{L(\w^{\mathrm{ag}}_n)} - L(\wopt) ~\le~  \tilde{\Ocal}\left(\sqrt{\frac{L(\wopt)}{bn}}+\frac{1}{\sqrt{b}n}+ \frac{1}{n^2}\right) ~=~ \tilde{\Ocal}\left(\sqrt{\frac{L(\wopt)}{m}}+\frac{\sqrt{b}}{m}+ \frac{b^2}{m^2}\right).
\end{equation}
To understand the implication these bounds, we follow the approach described in \cite{BotBou07,ShalSre08} to analyze large-scale learning algorithms. First, we fix a desired suboptimality parameter $\epsilon$, which measures how close to $L(\wopt)$ we want to get. Then, we assume that both algorithms are ran till the suboptimality of their outputs is at most $\epsilon$. Our goal would be to understand the \emph{runtime} each algorithm needs, till attaining suboptimality $\epsilon$, as a function of $L(\wopt),\epsilon,b$.

To measure this runtime, we need to discern two settings here: a \emph{parallel} setting, where we assume that the mini-batch gradient computations are performed in parallel, and a \emph{serial} setting, where the gradient computations are performed one after the other. In a parallel setting, we can take the number of iterations $n$ as a rough measure of the runtime (note that in both algorithms, the runtime of a single iteration is comparable). In a serial setting, the relevant parameter is $m$, the number of data accesses.

To analyze the dependence on $m$ and $n$, we upper bound
\eqref{eq:sgd} and \eqref{eq:ag} by $\epsilon$, and
invert\ignore{\footnote{The derivation is based on the fact that for
    any monotonically descreasing functions $f,g$, if we ignore
    constants, then a bound of the form $\epsilon\leq
    \Ocal(f(n)+g(n))$ is equivalent to $\epsilon
    \leq\Ocal(\max\{f(n),g(n)\})$, which in turn is equivalent to
    $n\leq \Ocal(\max\{f^{-1}(n),g^{-1}(n)\})$, or $n\leq
    \Ocal(f^{-1}(\epsilon)+g^{-1}(\epsilon))$. This also holds for
    sums of more functions.}} them to get the bounds on $m$ and $n$.
  Ignoring logarithmic factors, for the SGD algorithm we get
\begin{equation}\label{eq:sgdinv}
n\leq \frac{1}{\epsilon}\left(\frac{L(\wopt)}{\epsilon}\cdot\frac{1}{b}+1\right)\;\;\;\;\;\;
m\leq \frac{1}{\epsilon}\left(\frac{L(\wopt)}{\epsilon}+b\right),
\end{equation}
and for the AG algorithm we get
\begin{equation}\label{eq:aginv}
n\leq \frac{1}{\epsilon}\left(\frac{L(\wopt)}{\epsilon}\cdot\frac{1}{b}+\frac{1}{\sqrt{b}}+\sqrt{\epsilon}\right)\;\;\;\;\;\;
m\leq \frac{1}{\epsilon}\left(\frac{L(\wopt)}{\epsilon}+\sqrt{b}+b\sqrt{\epsilon}\right).
\end{equation}
First, let us compare the performance of these two algorithms in the parallel setting, where the relevant parameter to measure runtime is $n$. Analyzing which of the terms in each bound dominates, we get that for the SGD algorithm, there are 2 regimes, while for the AG algorithm, there are 2-3 regimes depending on the relationship between $L(\wopt)$ and $\epsilon$. The following two tables summarize the situation (again, ignoring constants):
\begin{center}
\renewcommand\arraystretch{1.5}
\begin{minipage}[b]{0.35\linewidth}\centering
SGD Algorithm
\begin{tabular}{|c|c|}\hline
Regime & n \\\hline\hline
$b\leq \sqrt{L(\wopt)m}$ & $\frac{L(\wopt)}{\epsilon^2 b}$\\
$b\geq \sqrt{L(\wopt)m}$ & $\frac{1}{\epsilon}$\\\hline
\end{tabular}
\end{minipage}
\hspace{0.5cm}
\begin{minipage}[b]{0.5\linewidth}
\centering
AG Algorithm
    \begin{tabular}{|c||c|c|}\hline
    & Regime & n \\\hline\hline
    \multirow{2}{*}{$\epsilon \leq L(\wopt)^2$}
    & $b\leq L(\wopt)^{1/4}m^{3/4}$ & $\frac{L(\wopt)}{\epsilon^2 b}$ \\
    & $b\geq L(\wopt)^{1/4}m^{3/4}$ & $\frac{1}{\sqrt{\epsilon}}$ \\\hline\hline
    \multirow{3}{*}{$\epsilon \geq L(\wopt)^2$}
    & $b\leq L(\wopt)m$ & $\frac{L(\wopt)}{\epsilon^2 b}$\\
    & $L(\wopt)m\leq b \leq m^{2/3}$ & $\frac{1}{\epsilon\sqrt{b}}$ \\
    & $b\geq m^{2/3}$ & $\frac{1}{\sqrt{\epsilon}}$\\\hline
    \end{tabular}
\end{minipage}
\end{center}
From the tables, we see that for both methods, there is an initial linear speedup as a function of the minibatch size $b$. However, in the AG algorithm, this linear speedup regime holds for much larger minibatch sizes\footnote{Since it is easily verified that $\sqrt{L(\wopt) m}$ is generally smaller than both $L(\wopt)^{1/4}m^{3/4}$ and $L(\wopt)m$}. Even beyond the linear speedup regime, the AG algorithm still maintains a $\sqrt{b}$ speedup, for the reasonable case where $\epsilon\geq L(\wopt)^2$. Finally, in all regimes, the runtime bound of the AG algorithm is equal or significantly smaller than that of the SGD algorithm.

We now turn to discuss the serial setting, where the runtime is measured in terms of $m$. Inspecting \eqref{eq:sgdinv} and \eqref{eq:aginv}, we see that a
larger size of $b$ actually requires $m$ to increase for both
algorithms. This is to be expected, since mini-batching does not lead
to large gains in a serial setting. However, using mini-batching in a
serial setting might still be beneficial for implementation reasons,
resulting in constant-factor improvements in runtime (e.g. saving
overhead and loop control, and via pipelining, concurrent memory
accesses etc.).  In that case, we can at least ask what is the largest
mini-batch size that won't degrade the runtime guarantee by more than
a constant.  Using our bounds, the mini-batch size $b$ for the SGD algorithm can scale as much as $L/\epsilon$, vs. a larger value of $L/\epsilon^{3/2}$ for the AG algorithm.

Finally, an interesting point is that the AG algorithm is sometimes actually
\emph{necessary} to obtain significant speed-ups via a mini-batch
framework (according to our bounds). Based on the table above, this happens when the desired suboptimality $\epsilon$ is not much bigger then $L(\wopt)$,
i.e.~$\epsilon=\Omega(L(\wopt))$.  This includes the ``separable''
case, $L(\wopt)=0$, and in general a regime where the ``estimation
error'' $\epsilon$ and ``approximation error'' $L(\wopt)$ are roughly
the same---an arguably very relevant one in machine learning. For the SGD algorithm, the critical mini-batch value $\sqrt{L(\wopt)m}$ can be shown to equal $L(\wopt)/\epsilon$, which is $O(1)$ in our case. So with SGD we get no non-constant parallel speedup. However, with AG, we still enjoy a speedup of at least $\Theta(\sqrt{b})$, all the way up to mini-batch size $b=m^{2/3}$.

\section{Experiments}

\begin{figure}[t]

\noindent \begin{centering}
\begin{tabular}{ @{} L @{} S @{} S @{} }
& astro-physics & CCAT \\
\rotatebox{90}{Test Loss} &
\includegraphics[width=0.44\textwidth]{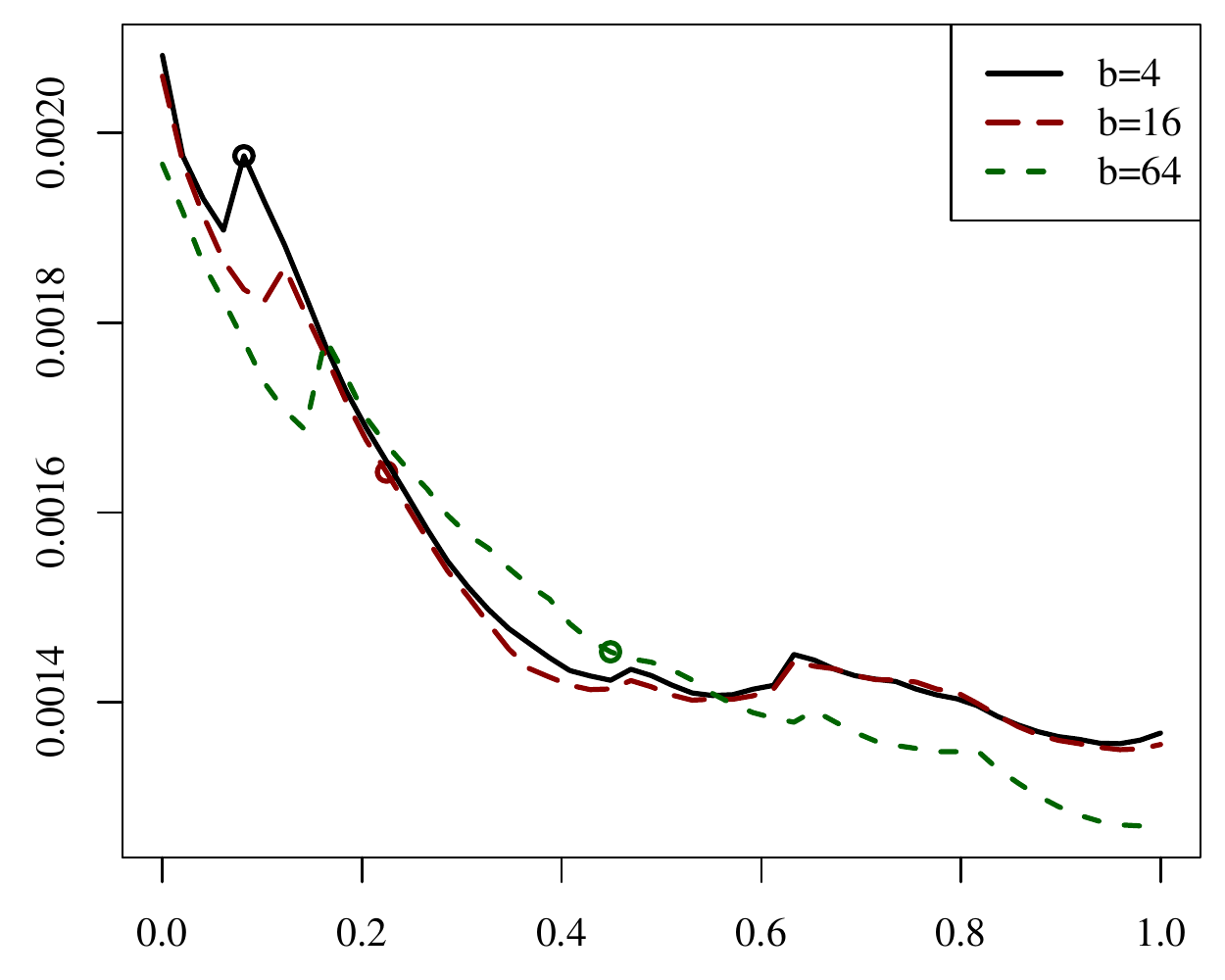} &
\includegraphics[width=0.44\textwidth]{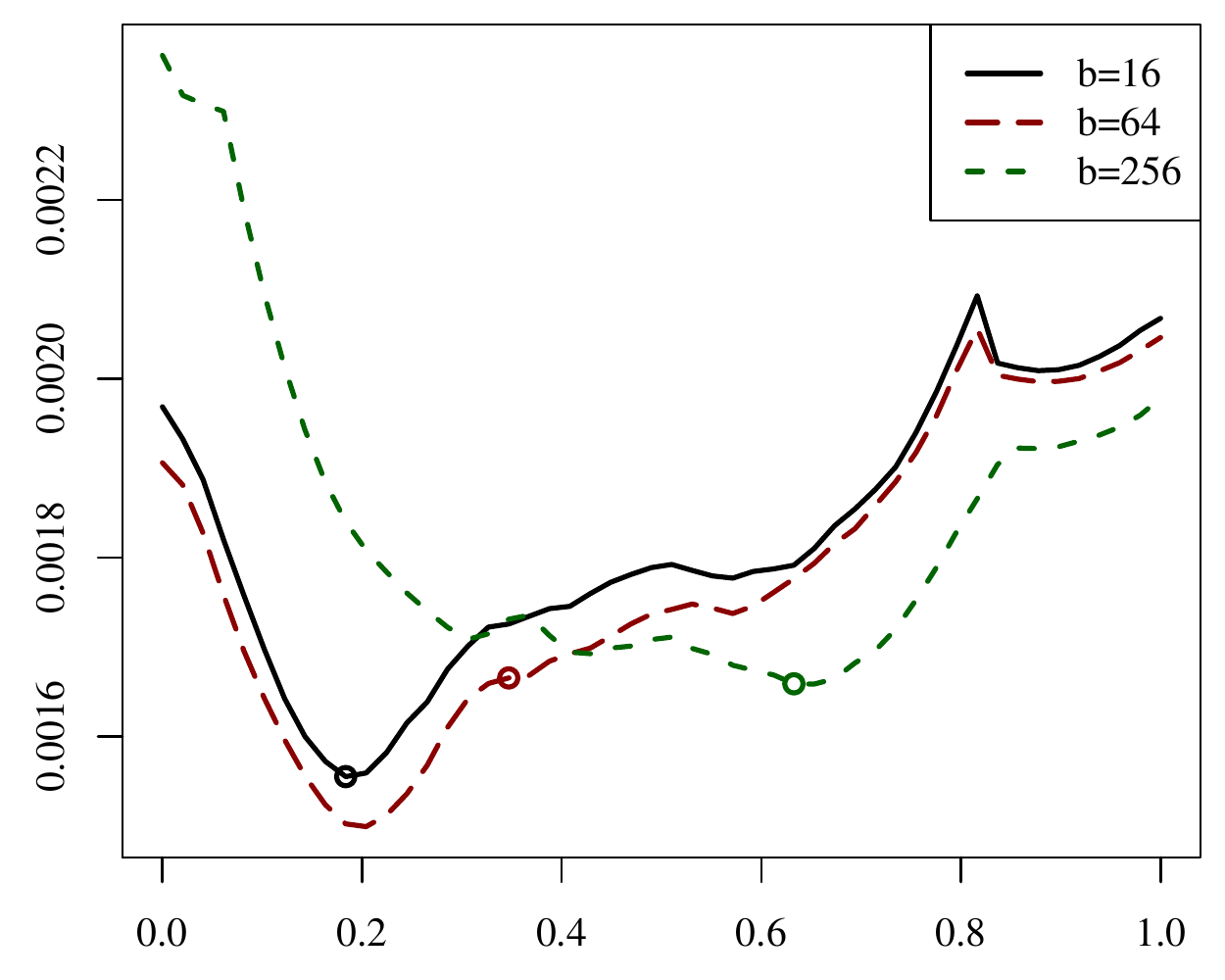} \\
& $p$ & $p$ \\
\end{tabular}
\end{centering}

{\small \caption{\small Left: Test smoothed hinge loss, as a function of $p$, after
training using the AG algorithm on 6361 examples from 
astro-physics, for various batch sizes. Right: the same, for 18578
examples from CCAT. In both datasets, margin violations were
removed before training so that $L(\wopt)=0$. The
circled points are the theoretically-derived values $p = \ln b / ( 2 \ln ( n -
1 ) )$ (see Theorem \ref{thm:ag}).}}

\label{fig:power}
\end{figure}

We implemented both the SGD algorithm (Algorithm \ref{alg:md})
and the AG algorithm (Algorithm \ref{alg:ag}, using step-sizes of the form $\gamma_i=\gamma i^p$ as suggested by \thmref{thm:ag}) on two publicly-available binary classification problems, astro-physics and CCAT. We used the smoothed hinge loss $\ell(\w;\x,y)$, defined as $0.5-y\w^{\top}\x$ if $y\w^{\top}\x\leq 0$; $0$ if $y\w^{\top}\x> 1$, and $0.5(1-y\w^{\top}\x)^2$ in between.

While both datasets are relatively easy to classify, we also wished to understand the algorithms' performance in the ``separable'' case $L(\wopt)=0$, to see if the theory in \secref{sec:minibatches} holds in practice. To this end, we created an additional version of each dataset, where $L(\wopt)=0$, by training a classifier on the entire dataset and removing margin violations.

In all of our experiments, we used up to half of the data for training, and
one-quarter each for validation and testing. The validation set was used to
determine the step sizes $\eta$ and $\gamma_i$. We justify this by noting that our goal is to compare the performance of the SGD and AG algorithms, independently of the difficulties in choosing their stepsizes. In the implementation, we neglected the projection step, as we found it does not significantly affect performance when the stepsizes are properly selected.

In our first set of experiments, we attempted to determine the relationship
between the performance of the AG algorithm and the $p$
parameter, which determines the rate of increase of the step sizes $\gamma_i$.
Our experiments are summarized in Figure \ref{fig:power}. Perhaps the most
important conclusion to draw from these plots is that neither the
``traditional'' choice $p=1$, nor the constant-step-size choice $p=0$, give the best performance in all circumstances. Instead, there is a complicated
data-dependent relationship between $p$, and the final classifier's
performance. Furthermore, there appears to be a weak trend towards higher $p$
performing better for larger minibatch sizes $b$, which corresponds neatly with our theoretical predictions.

\begin{figure}[h!]

\noindent \begin{centering}
\begin{tabular}{ @{} L @{} S @{} S @{} }
& astro-physics & CCAT \\
\rotatebox{90}{Test Loss} &
\includegraphics[width=0.4\textwidth]{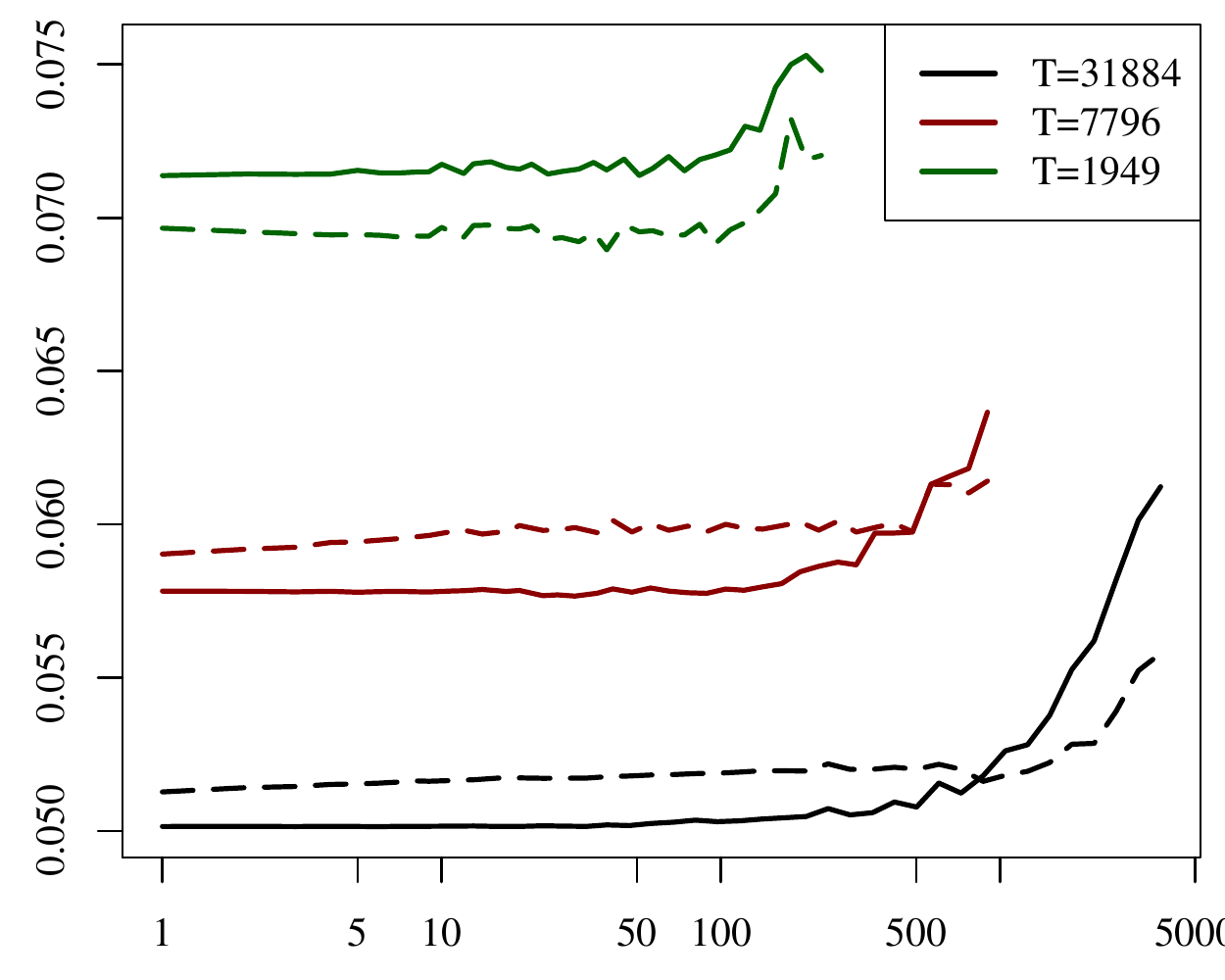} &
\includegraphics[width=0.4\textwidth]{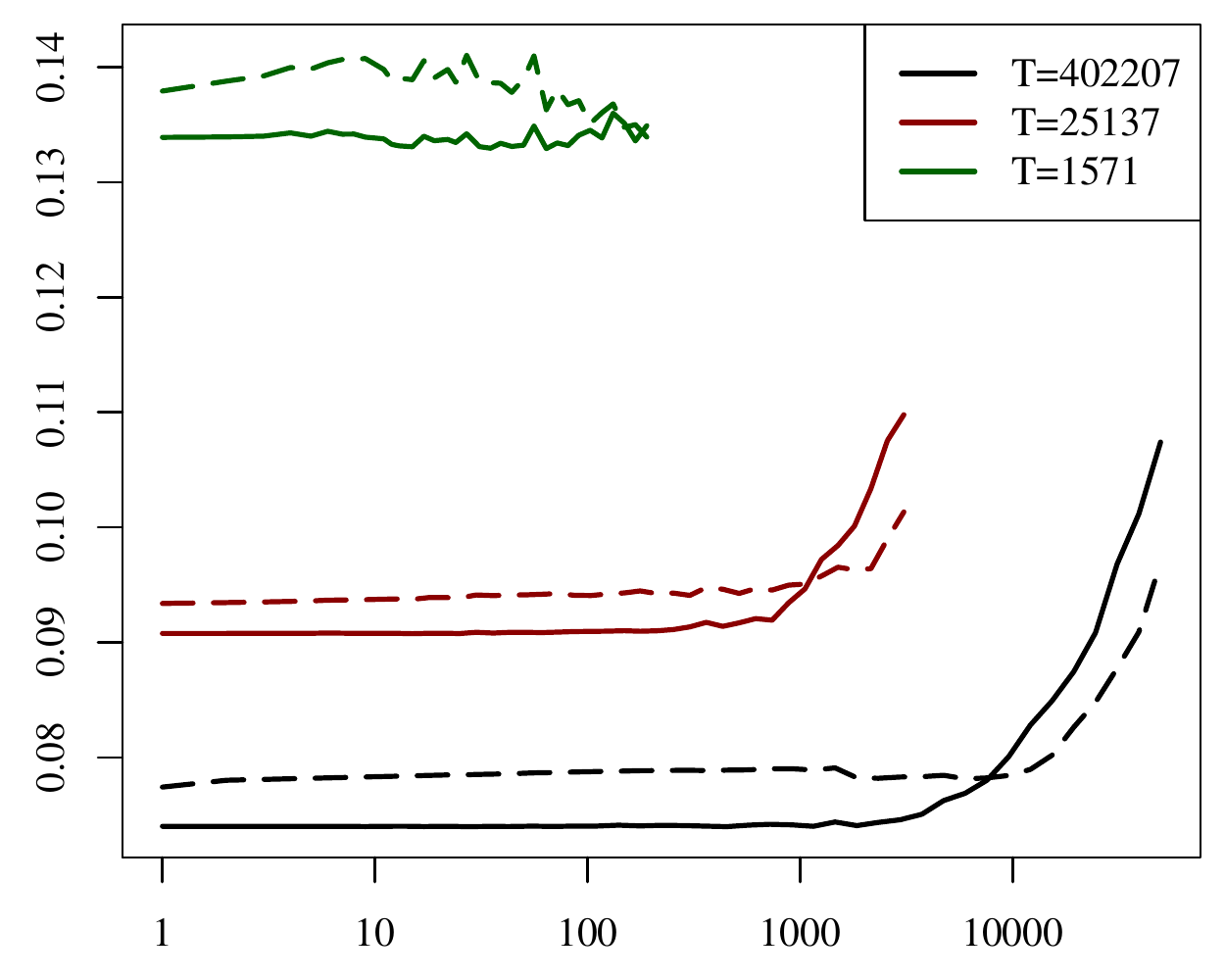} \\
\rotatebox{90}{Test Loss} &
\includegraphics[width=0.4\textwidth]{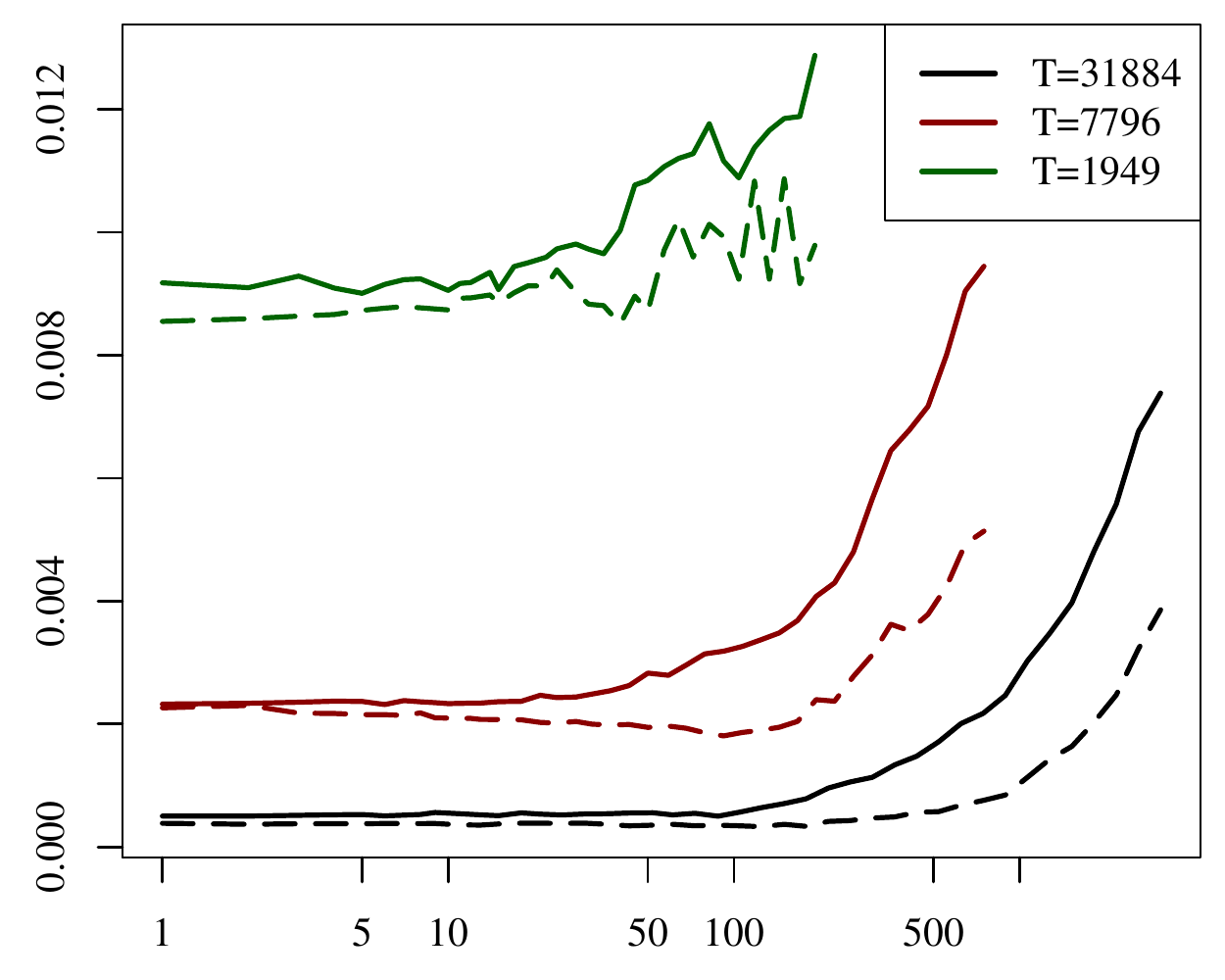} &
\includegraphics[width=0.4\textwidth]{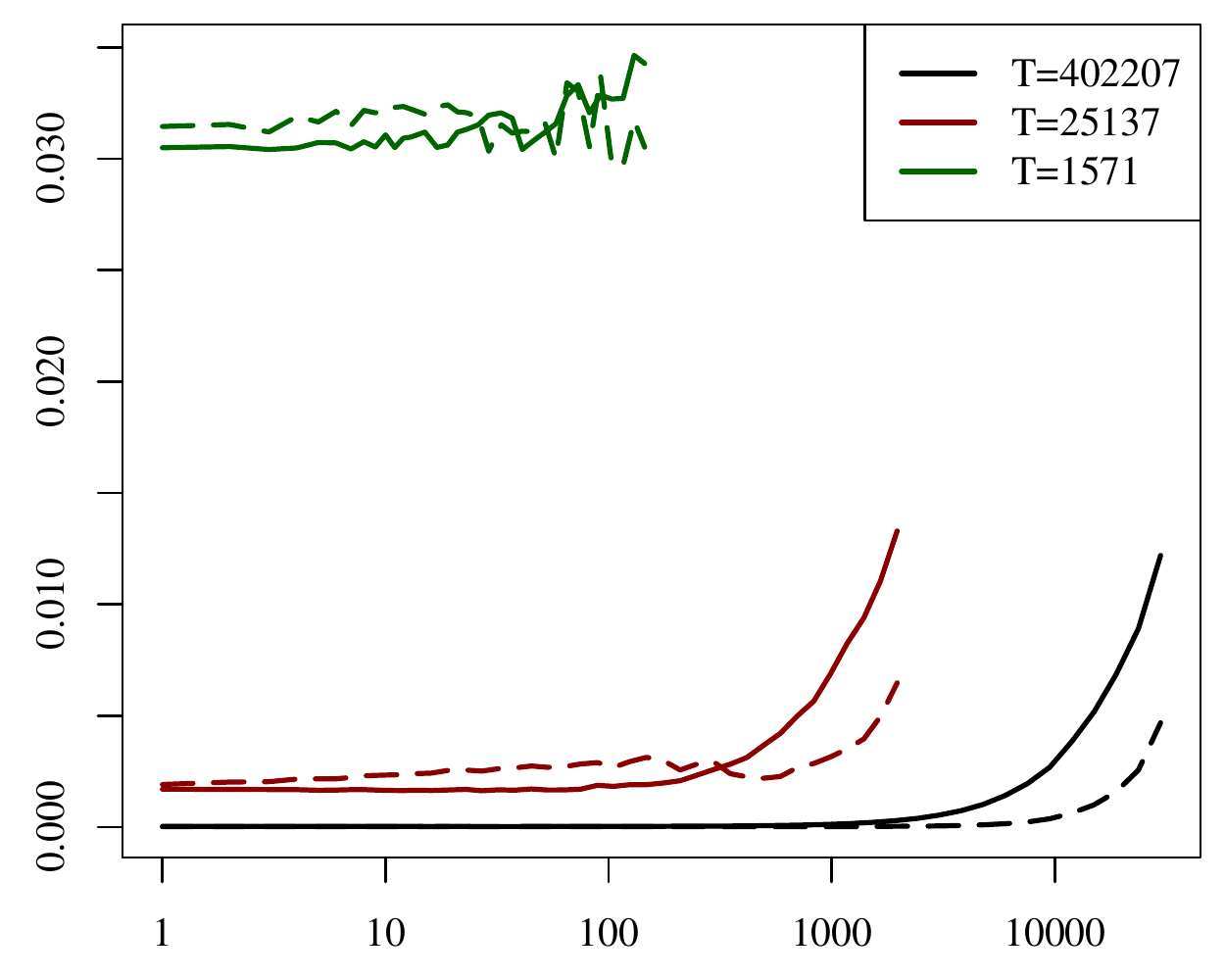} \\
\rotatebox{90}{Test Misclassification} &
\includegraphics[width=0.4\textwidth]{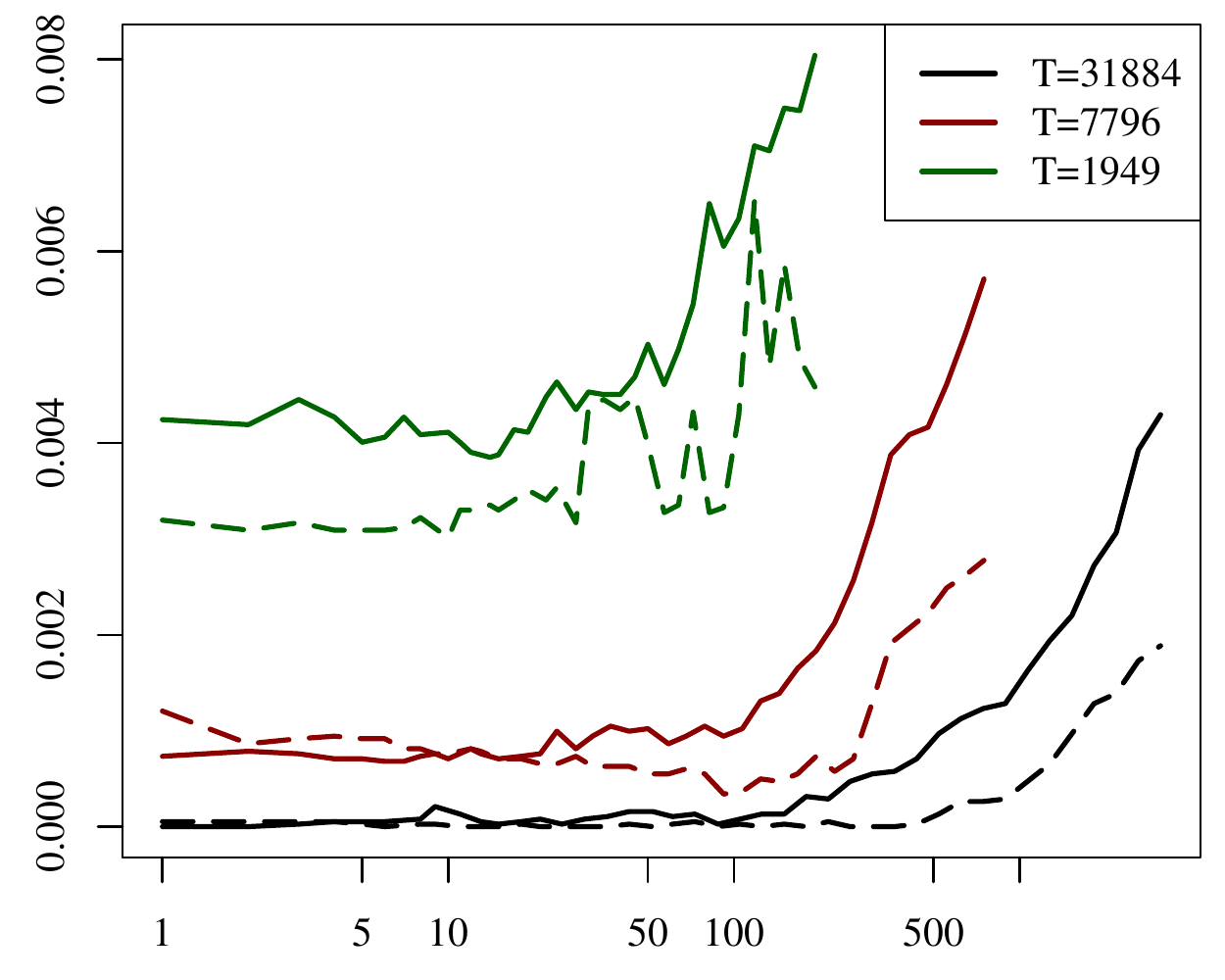} &
\includegraphics[width=0.4\textwidth]{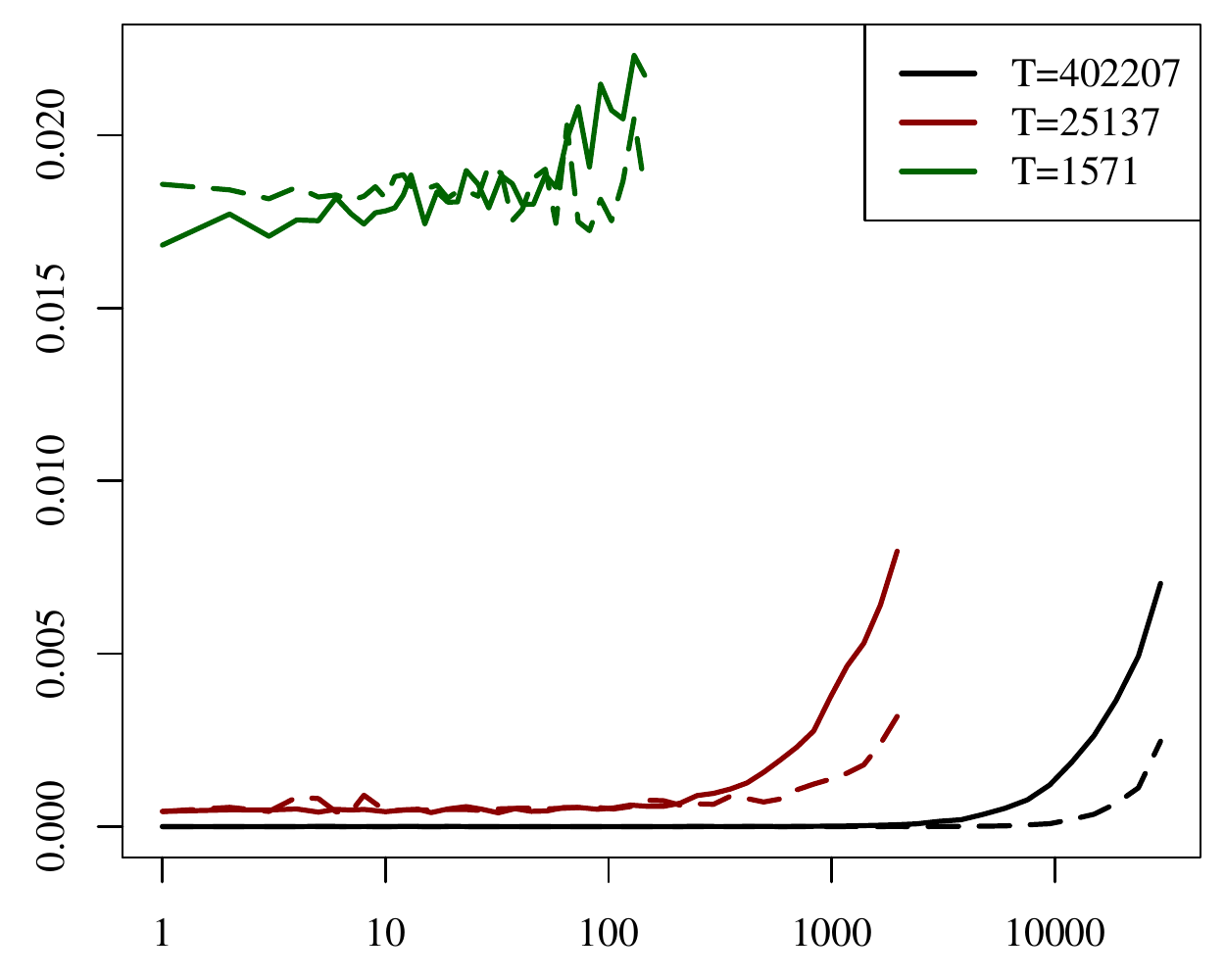} \\
& $b$ & $b$ \\
\end{tabular}
\end{centering}

{\small \caption{\small Test loss on astro-physics and CCAT as a function of
mini-batch size $b$ (in log-scale), where the total amount of training data
$m=nb$ is held fixed. Solid lines and dashed lines are for SGD and AG
respectively (for AG, we used $p = \ln b / ( 2 \ln( n - 1 ) )$ as in Theorem
\ref{thm:ag}). The upper row shows the smoothed hinge loss on the test set,
using the original (uncensored) data.  The bottom rows show the smoothed hinge
loss and misclassification rate on the test set, using the modified data where
$L(\wopt)=0$. All curves are averaged over three runs.}}

\label{fig:batch}
\end{figure}

In our next experiment, we directly compared the performance of the
SGD and AG methods. To do so, we varied the minibatch size $b$ while
holding the total amount of data used for training, $m=nb$, fixed.
When $L(\wopt)>0$ (top row of Figure \ref{fig:batch}), the total
sample size $m$ is high and the suboptimality $\epsilon$ is low (red
and black plots), we see that for small minibatch size, both methods
do not degrade as we increase $b$, corresponding to a linear parallel
speedup. In fact, SGD is actually overall better, but as $b$
increases, its performance degrades more quickly, eventually
performing worse than AG.  That is, even in the least favorable
scenario for AG (high $L(\wopt)$ and small $\epsilon$, see the tables
in \secref{sec:minibatches}), it does give benefits with large enough
minibatch sizes. Also, we see that even here, once the suboptimality
$\epsilon$ is roughly equal to $L(\wopt)$, AG significantly
outperforms SGD, even with small minibatches, agreeing with our the
theory.

Turning to the case $L(\wopt)=0$ (bottom two rows of Figure
\ref{fig:batch}), which is theoretically more favorable to AG, we see
it is indeed mostly better, in terms of retaining linear parallel
speedups for larger minibatch sizes, even for large data set sizes
corresponding to small suboptimality values, and might even be
advantageous with small minibatch sizes.

\section{Summary}

In this paper, we presented novel contributions to the theory of
first order stochastic convex optimization (Theorems \ref{thm:sgd} and
\ref{thm:ag}, generalizing results of \cite{DMB} and \cite{Lan09}
to be sensitive to $\L{\wopt}$), developed a novel step size
strategy for the accelerated method that we used in order to obtain
our results and we saw works well in practice, and provided a more
refined analysis of the effects of minibatching which paints a
different picture then previous analyses \cite{DMB,AD} and highlights
the benefit of accelerated methods.

A remaining open practical and theoretical question is whether the
bound of Theorem \ref{thm:ag} is tight.  Following \cite{Lan09}, the
bound is tight for $b=1$ and $b\rightarrow\infty$, i.e.~the first and
third terms are tight, but it is not clear whether the $1/(\sqrt{b}n)$
dependence is indeed necessary.  It would be interesting to understand
whether with a more refined analysis, or perhaps different step-sizes,
we can avoid this term, whether an altogether different algorithm is
needed, or whether this term does represent the optimal behavior for
any method based on $b$-aggregated stochastic gradient estimates.


\bibliographystyle{plain}
\bibliography{mybib}

\begin{thebibliography}{10}

\bibitem{AD}
A.~Agarwal and J.~Duchi.
\newblock Distributed delayed stochastic optimization.
\newblock Technical report, arXiv, 2011.

\bibitem{BeckTeb03}
A.~Beck and M.~Teboulle.
\newblock Mirror descent and nonlinear projected subgradient methods for convex
  optimization.
\newblock {\em Operations Research Letters}, 31(3):167 -- 175, 2003.

\bibitem{BotBou07}
L.~Bottou and O.~Bousquet.
\newblock The tradeoffs of large scale learning.
\newblock In {\em NIPS}, 2007.

\bibitem{DMB}
O.~Dekel, R.~Gilad Bachrach, O.~Shamir, and L.~Xiao.
\newblock Optimal distributed online prediction using mini-batches.
\newblock Technical report, arXiv, 2010.

\bibitem{Lan09}
G.~Lan.
\newblock An optimal method for stochastic convex optimization.
\newblock Technical report, Georgia Institute of Technology, 2009.

\bibitem{NemirovskiJuLaSh09}
A.~Nemirovski, A.~Juditsky, G.~Lan, and A.~Shapiro.
\newblock Robust stochastic approximation approach to stochastic programming.
\newblock {\em SIAM Journal on Optimization}, 19(4):1574--1609, 2009.

\bibitem{Nesterov83}
Y.~Nesterov.
\newblock A method for unconstrained convex minimization problem with the rate
  of convergence $o(1/k^2)$.
\newblock {\em Doklady AN SSSR}, 269:543--547, 1983.

\bibitem{Nesterov05}
Y.~Nesterov.
\newblock Smooth minimization of non-smooth functions.
\newblock {\em Math. Program.}, 103(1):127--152, 2005.

\bibitem{Nesterov09}
Y.~Nesterov.
\newblock Primal-dual subgradient methods for convex problems.
\newblock {\em Mathematical Programming}, 120(1):221--259, August 2009.

\bibitem{DekelGilShamXia11}
O.~Shamir O.~Dekel, R. Gilad-Bachrach and L.~Xiao.
\newblock Optimal distributed online prediction.
\newblock In {\em ICML}, 2011.

\bibitem{ShalSiSreCo11}
S.~Shalev-Shwartz, Y.~Singer, N.~Srebro, and A.~Cotter.
\newblock Pegasos: primal estimated sub-gradient solver for {SVM}.
\newblock {\em Math. Program.}, 127(1):3--30, 2011.

\bibitem{ShalSre08}
S.~Shalev-Shwartz and N.~Srebro.
\newblock {SVM} optimization: inverse dependence on training set size.
\newblock In {\em ICML}, 2008.

\bibitem{SreSriTew10}
N.~Srebro, K.~Sridharan, and A.~Tewari.
\newblock Smoothness, low noise and fast rates.
\newblock In {\em NIPS}, 2010.

\bibitem{Shalev07}
S.Shalev-Shwartz.
\newblock {\em Online Learning: Theory, Algorithms, and Applications}.
\newblock PhD thesis, Hebrew University of Jerusalem, 2007.

\bibitem{Tseng08}
P.~Tseng.
\newblock On accelerated proximal gradient methods for convex-concave
  optimization.
\newblock {\em Submitted to SIAM Journal on Optimization}, 2008.

\bibitem{Xiao10}
L.~Xiao.
\newblock Dual averaging methods for regularized stochastic learning and online
  optimization.
\newblock {\em Journal of Machine Learning Research}, 11:2543--2596, 2010.

\end{thebibliography}

\newpage

\appendix

\section{Generalizing to Different Norms}

We now turn to general norms and discuss the generic Mirror Descent
and Accelerated Mirror Descent algorithms.  In this more general case
we let domain $\Wcal$ be some closed convex set of a Banach space
equipped with norm $\norm{\cdot}$. We will use $\norm{\cdot}_*$ to represent the dual norm of $\norm{\cdot}$. Further the $H$-smoothness of the loss function in this general case is takes the form that for any $z \in \Z$ and any $\w , \w' \in \Wcal$,
$$
\norm{\nabla \ell(\w , z) - \nabla \ell(\w',z)}_* \le H \norm{\w - \w'}
$$
The key to generalizing the algorithms and result is to find a non-negative function
$R : \Wcal \mapsto \mathbb{R}$ that is strongly convex on the domain $\Wcal$ w.r.t. to the norm $\norm{\cdot}$, that is:
\begin{definition}
A function $R : \Wcal \mapsto \reals$ is said to be $1$-strongly convex w.r.t. norm $\norm{\cdot}$ if for any $\w , \w' \in \Wcal$ and any $\alpha \in [0,1]$,
\begin{align*}
R(\alpha \w + (1 - \alpha)\w' ) \le \alpha R(\w) + (1 - \alpha) R(\w') - \tfrac{\alpha (1 - \alpha)}{2} \norm{\w - \w'}^2
\end{align*}
\end{definition}
We also denote more generally 
$$D :=  \sqrt{2 \sup_{\w \in \Wcal} R(\w)}~.$$ 
The generalizations of the SGD and AG methods are summarized in
Algorithms \ref{alg:smd} and \ref{alg:amd} respectively. The key
difference between these and the Euclidean case is that the gradient
descent step is replaced by a descent step involving gradient mappings
of $R$ and its conjugate $R^*$ and the projection step is replaced by
Bregman projection (projection to set minimizing the Bregman
divergence to the point).

\begin{algorithm}
\caption{Stochastic Mirror Descent with Mini-Batching (SMD)}
\begin{algorithmic}
\STATE Parameters: Step size $\eta$, mini-batch size $b$.
\STATE Input: Sample $\z_1,\ldots,\z_m$
\STATE $\w_1= \argmin{\w} R(\w)$
\FOR{$i=1$ to $n=m/b$}
    \STATE Let $\ell_i(\w_i) = \frac{1}{b}\sum_{t=b(i-1)+1}^{bi}\ell(\w_i,\z_t)$
    \STATE \hspace{-0.08in}$\left.\begin{array}{ll}
\w'_{i+1} :=  \nabla R^*\left( \nabla R\left(\w_{i}\right) - \gamma_i \nabla \ell_i(\w_i) \right) \\
\w_{i+1} :=  \argmin{\w \in \Wcal}\breg{R}{\w}{\w'_{i+1}}
\end{array}\right\} \w_{i+1} = \argmin{\w \in \Wcal}\left\{\eta \ip{\nabla \ell_i(\w_i)}{\w - \w_i} + \breg{R}{\w}{\w_i} \right\}
$
\ENDFOR
\STATE Return $\bar{\w} = \frac{1}{n}\sum_{i=1}^{n}\w_i$
\end{algorithmic}
\label{alg:smd}
\end{algorithm}

\begin{algorithm}
\caption{Accelerated Mirror Descent Method (AMD)}
\begin{algorithmic}
\STATE Parameters: Step sizes $(\gamma_i,\beta_i)$, mini-batch size $b$
\STATE Input: Sample $\z_1,\ldots,\z_m$
\STATE $\w_1 = \argmin{\w} R(\w)$
\FOR{$i=1$ to $n=m/b$}
    \STATE Let $\ell_i(\w_i) := \frac{1}{b}\sum_{t=b(i-1)+1}^{bi}\ell(\w,\z_t)$
    \STATE $\w^{\mathrm{md}}_i := \beta_i^{-1} \w_{i} + (1 - \beta_i^{-1}) \w_{i}^{\mathrm{ag}}$
    \STATE \hspace{-0.1in} $\left.\begin{array}{ll}
\w'_{i+1} :=  \nabla R^*\left( \nabla R\left(\w^{\mathrm{md}}_{i}\right) - \gamma_i \nabla \ell_i(\w^{\mathrm{md}}_i) \right) \\
\w_{i+1} :=  \argmin{\w \in \Wcal}\breg{R}{\w}{\w'_{i+1}}
\end{array}\right\} \w_{i+1} = \argmin{\w \in \Wcal}\left\{\gamma_i \ip{\nabla \ell_i(\w^{\mathrm{md}}_i)}{\w - \w^{\mathrm{md}}_i} + \breg{R}{\w}{\w^{\mathrm{md}}_i} \right\}
$
\STATE $\w^{\mathrm{ag}}_{i+1} \leftarrow \beta_i^{-1} \w_{i+1} + (1 - \beta_i^{-1}) \w_{i}^{\mathrm{ag}}$
\ENDFOR
\STATE Return $\w^{\mathrm{ag}}_{n}$
\end{algorithmic}
\label{alg:amd}
\end{algorithm}

\begin{theorem}\label{thm:smd}
Let $R : \Wcal \mapsto \reals$ be  a non-negative strongly convex
function on $\Wcal$ w.r.t. norm $\norm{\cdot}$. Let $K = \sqrt{2
  \sup_{\w : \norm{\w} \le 1} R(\w)}$. For any $\wopt\in\Wcal$, using
Stochastic Mirror Descent with a step size of
$$
\eta = \min\left\{\frac{1}{2 H}, \frac{b}{32 H K^2 }, \frac{\sqrt{\frac{32 b R(\wopt)}{ L(\wopt) H K^2  n}}}{16\left(1 + \sqrt{\frac{32 H K^2  R(\wopt)}{ L(\wopt) b n}}\right) } \right\} ~ ,
$$
we have that,
\begin{align*}
\E{L(\bar{\w})} - L(\wopt) \le \sqrt{\frac{128 H K^2  R(\wopt)\ L(\wopt)}{b n}} +  \frac{4 L(\wopt) + 8 H R(\wopt)}{n} + \frac{16 H K^2  R(\wopt)}{b n}
\end{align*}
\end{theorem}

\begin{theorem}\label{thm:amd}
Let $R : \Wcal \mapsto \reals$ be a non-negative strongly convex
function on $\Wcal$ w.r.t. norm $\norm{\cdot}$. Also let $K = \sqrt{2
  \sup_{\w : \norm{\w} \le 1} R(\w)}$.  For any $\wopt\in\Wcal$, using
Accelerated Mirror Descent with step size parameters $\beta_i = \frac{i+1}{2}$, $\gamma_i = \gamma i^p$ where 
$$
\gamma = \min \left\{\frac{1}{4 H},\ \sqrt{\frac{b R(\wopt)}{174 H K^2 L(\wopt) (n-1)^{2p+1}}},\  \left(\frac{b}{1044 H K^2 (n - 1)^{2p}}\right)^{\frac{p+1}{2p+1}} \left(\frac{6 R(\wopt) }{ \frac{3}{2} H D^2 + L(\wopt)} \right)^{\frac{p}{2p+1}}\right\} ~~\textrm{and}\\
$$
$$
p = \min\left\{\max\left\{\frac{\log(b)}{2 \log(n-1)} , \frac{\log \log(n)}{2\left(\log(b(n-1)) - \log \log(n)\right)}\right\} ,1\right\}~~,
$$
as long as $ n \ge \max\{783 K^2, \frac{87 K^2 L(\wopt)}{HD^2}\}$, we have that :
\begin{align*}
\E{L(\w^{\mathrm{ag}}_n)} - L(\wopt) & \le 164 \sqrt{\frac{H K^2 R(\wopt) L(\wopt)}{b (n-1)}} + \frac{580  H K^2  (R(\wopt))^{2/3}  D^{\frac{2}{3}} }{\sqrt{b} (n-1)}   +  \frac{545 H K^2 D^2 \sqrt{\log(n)}}{b (n - 1)}  + \frac{8 H R(\wopt) }{ (n - 1)^{2}} 
\end{align*}
\end{theorem}

\section{Complete Proofs}

We provide complete proofs of Theorems \ref{thm:smd} and
\ref{thm:amd}, noting how Theorems \ref{thm:sgd} and \ref{thm:ag} are
specializations to the Euclidean case.

\subsection{Stochastic Mirror Descent}
\begin{proof}[Proof of Theorem \ref{thm:smd}]
Due to $H$-smoothness of convex function $L$ we have that,
\begin{align*}
L(\w_{i+1}) & \le L(\w_i) + \ip{\nabla L(\w_i)}{\w_{i+1} - \w_i} + \frac{H}{2} \|\w_{i+1} - \w_{i}\|^2 \\
& = L(\w_i) + \ip{\nabla L(\w_i) - \nabla \ell_i(\w_i)}{\w_{i+1} - \w_i} + \frac{H}{2} \|\w_{i+1} - \w_{i}\|^2 + \ip{\nabla \ell_i(\w_i)}{\w_{i+1} - \w_i}\\
\intertext{by Holder's inequality we get,}
& \le L(\w_i) + \|\nabla L(\w_i) - \nabla \ell_i(\w_i)\|_* \|\w_{i+1} - \w_i\| + \frac{H}{2} \|\w_{i+1} - \w_{i}\|^2 + \ip{\nabla \ell_i(\w_i)}{\w_{i+1} - \w_i}\\
\intertext{since for any $\alpha > 0$, $ab \le \frac{a^2}{2 \alpha} + \frac{\alpha b^2}{2}$,}
& \le L(\w_i) + \frac{\|\nabla L(\w_i) - \nabla \ell_i(\w_i)\|_*^2 }{2(1/\eta - H)} + \frac{(1/\eta - H)}{2} \|\w_{i+1} - \w_i\|^2 + \frac{H}{2} \|\w_{i+1} - \w_{i}\|^2 + \ip{\nabla \ell_i(\w_i)}{\w_{i+1} - \w_i}\\
& = L(\w_i) + \frac{\|\nabla L(\w_i) - \nabla \ell_i(\w_i)\|_*^2 }{2(1/\eta - H)} + \frac{\|\w_{i+1} - \w_i\|^2}{2 \eta}   + \ip{\nabla \ell_i(\w_i)}{\w_{i+1} - \w_i}
\end{align*}
We now note that the update step can be written equivalently as 
$$\w_{i+1} = \argmin{\w \in \Wcal} \left\{\eta \ip{\nabla \ell_i(\w_i)}{\w - \w_i} + \Delta_R(\w , \w_i)\right\}~.$$ 
It can be shown that (see for instance Lemma 1 of \cite{Lan09}) 
$$
\eta \ip{\nabla \ell_i(\w_i)}{\w_{i+1} - \w_i} \le \eta \ip{\nabla \ell_i(\w_i)}{\wopt - \w_i} + \Delta_R(\wopt, \w_{i}) - \Delta_R(\wopt , \w_{i+1}) - \Delta_R(\w_i , \w_{i+1})
$$
Plugging this we get that,
\begin{align*}
L(\w_{i+1}) &\le L(\w_i) + \frac{\|\nabla L(\w_i) - \nabla \ell_i(\w_i)\|_*^2}{2 (1/\eta - H)} + \frac{\norm{\w_{i} - \w_{i+1}}^2}{2\eta} + \ip{\nabla \ell_i(\w_i)}{\wopt - \w_i}\\
&~~~~~~~~~~~~~ + \frac{1}{\eta}\left(\Delta_R(\wopt, \w_{i}) - \Delta_R(\wopt , \w_{i+1}) - \Delta_R(\w_i , \w_{i+1})\right) \\
& = L(\w_i) + \frac{\|\nabla L(\w_i) - \nabla \ell_i(\w_i)\|_*^2}{2 (1/\eta - H)} + \frac{\norm{\w_{i} - \w_{i+1}}^2}{2\eta} + \ip{\nabla \ell_i(\w_i) - \nabla L(\w_i)}{\wopt - \w_i} + \ip{\nabla L(\w_i)}{\wopt - \w_i}\\
&~~~~~~~~~~~~~ + \frac{1}{\eta}\left( \Delta_R(\wopt, \w_{i}) - \Delta_R(\wopt , \w_{i+1}) - \Delta_R(\w_i , \w_{i+1}) \right) \\
& \ge L(\w_i) + \frac{\|\nabla L(\w_i) - \nabla \ell_i(\w_i)\|_*^2}{2 (1/\eta - H)} + \frac{\norm{\w_{i} - \w_{i+1}}^2}{2\eta} + \ip{\nabla \ell_i(\w_i) - \nabla L(\w_i)}{\wopt - \w_i} - \ip{\nabla L(\w_i)}{\w_i - \wopt}\\
&~~~~~~~~~~~~~ + \frac{1}{\eta}\left( \Delta_R(\wopt, \w_{i}) - \Delta_R(\wopt , \w_{i+1}) - \Delta_R(\w_i , \w_{i+1}) \right) 
\intertext{by strong convexity, $\Delta_R(\w_i , \w_{i+1}) \ge \frac{\norm{\w_i - \w_{i+1}}}{2}$ and so, }
&\le L(\w_i) + \frac{\|\nabla L(\w_i) - \nabla \ell_i(\w_i)\|_*^2}{2 (1/\eta - H)} + \ip{\nabla \ell_i(\w_i) - \nabla L(\w_i)}{\wopt - \w_i} - \ip{\nabla L(\w_i)}{\w_i - \wopt}\\
&~~~~~~~~~~~~~ + \frac{1}{2 \eta}\left(\Delta_R(\wopt, \w_{i}) - \Delta_R(\wopt , \w_{i+1})\right) 
\intertext{since $\eta \le \frac{1}{2 H}$,}
& \le L(\w_i) + \eta \|\nabla L(\w_i) - \nabla \ell_i(\w_i)\|_*^2 + \ip{\nabla \ell_i(\w_i) - \nabla L(\w_i)}{\wopt - \w_i} - \ip{\nabla L(\w_i)}{\w_i - \wopt}\\
&~~~~~~~~~~~~~ + \frac{1}{\eta}\left( \Delta_R(\wopt, \w_{i}) - \Delta_R(\wopt , \w_{i+1})\right) 
\intertext{by convexity, $L(\w_{i}) - \ip{\nabla L(\w_i)}{\w_i - \wopt} \le L(\wopt)$ and so}
& \le L(\wopt) + \eta \|\nabla L(\w_i) - \nabla \ell_i(\w_i)\|_*^2 + \ip{\nabla \ell_i(\w_i) - \nabla L(\w_i)}{\wopt - \w_i}\\
&~~~~~~~~~~~~~ + \frac{1}{\eta}\left( \Delta_R(\wopt, \w_{i}) - \Delta_R(\wopt , \w_{i+1})\right) 
\end{align*}

Hence we conclude that :
\begin{align*}
\frac{1}{n-1} \sum_{i=1}^{n-1} L(\w_{i+1}) - L(\wopt) & \le \frac{\eta}{ (n-1) } \sum_{i=1}^{n-1}  \|\nabla L(\w_i) - \nabla \ell_i(\w_i)\|_*^2  + \frac{1}{n-1} \sum_{i=1}^{n-1} \ip{\nabla \ell_i(\w_i) - \nabla L(\w_i)}{\wopt - \w_i} \\
&~~~~~~~~~~ + \frac{1}{n-1} \sum_{i=1}^{n-1}\frac{\Delta_R(\wopt, \w_{i}) - \Delta_R(\wopt , \w_{i+1})}{\eta}\\
& = \frac{\eta}{ (n-1) } \sum_{i=1}^{n-1}  \|\nabla L(\w_i) - \nabla \ell_i(\w_i)\|_*^2  + \frac{1}{n-1} \sum_{i=1}^{n-1} \ip{\nabla \ell_i(\w_i) - \nabla L(\w_i)}{\wopt - \w_i} \\
&~~~~~~~~~~ + \frac{\breg{R}{\wopt}{\w_1} - \breg{R}{\wopt}{\w_{n-1}}}{\eta (n-1)} \\
& \le \frac{\eta}{ (n-1) } \sum_{i=1}^{n-1}  \|\nabla L(\w_i) - \nabla \ell_i(\w_i)\|_*^2  + \frac{1}{n-1} \sum_{i=1}^{n-1} \ip{\nabla \ell_i(\w_i) - \nabla L(\w_i)}{\wopt - \w_i} \\
&~~~~~~~~~~ + \frac{R(\wopt)}{\eta (n-1)} \\
& \le \frac{\eta}{ (n-1) } \sum_{i=1}^{n-1}  \|\nabla L(\w_i) - \nabla \ell_i(\w_i)\|_*^2  + \frac{1}{n-1} \sum_{i=1}^{n-1} \ip{\nabla \ell_i(\w_i) - \nabla L(\w_i)}{\wopt - \w_i} \\
&~~~~~~~~~~ + \frac{R(\wopt)}{\eta (n-1)} 
\end{align*}
Taking expectation with respect to sample on both sides and noticing that
 $\E{\ip{\nabla \ell_i(\w_i) - \nabla L(\w_i)}{\wopt - \w_i}} = 0$, we get that,
\begin{align*}
\E{\frac{1}{n-1} \sum_{i=1}^{n-1} L(\w_{i+1}) - L(\wopt)} & \le \frac{\eta}{ (n-1) } \sum_{i=1}^{n-1}  \E{\|\nabla L(\w_i) - \nabla \ell_i(\w_i)\|_*^2 } + \frac{R(\wopt)}{\eta (n-1)} 
\end{align*}
Now note that 
$$
\nabla L(\w_i) - \nabla \ell_i(\w_i) = \frac{1}{b} \sum_{t= (i-1)b+ 1}^{bi} \left(\nabla L(\w_i) - \nabla \ell(\w_i,z_t)\right)
$$
and that $\left(\nabla L(\w_i) - \ell(\w_i,z_t)\right)$ is a mean zero vector drawn i.i.d.  Also note that $\w_i$ only depends on the first $(i-1)b$ examples and so when we consider expectation w.r.t. $z_{(i-1)b + 1} , \ldots, z_{ib}$ alone, $\w_i$ is fixed.  Hence by Corollary \ref{cor:eqsmooth} we have that,
\begin{align*}
\E{\norm{\nabla L(\w_i) - \nabla \ell_i(\w_i)}_*^2} & \le \frac{K^2}{b^2}\ \E{\norm{\sum_{t= (i-1)b+ 1}^{bi} \left(\nabla L(\w_i) - \nabla  \ell(\w_i,z_t)\right)}_*^2}\\
& = \frac{K^2}{b^2} \sum_{t= (i-1)b+ 1}^{bi} \E{\norm{ \left(\nabla L(\w_i) - \nabla \ell(\w_i,z_t)\right)}_*^2}
\end{align*}
Plugging this back we get that
\begin{align*}
\E{\frac{1}{n-1} \sum_{i=1}^{n-1} L(\w_{i+1}) - L(\wopt)} & \le  \frac{K^2 \eta}{ b^2 (n-1) } \sum_{i=1}^{n-1}  \sum_{t= (i-1)b+ 1}^{bi} \E{\norm{ \left(\nabla L(\w_i) - \nabla \ell(\w_i,z_t)\right)}_*^2} + \frac{R(\wopt)}{\eta (n-1)} \\
& \le  \frac{2 K^2 \eta }{b^2 (n-1)  } \sum_{i=1}^{n-1} \sum_{t=(i-1)b + 1}^{ib}  \E{ \norm{\nabla L(\w_i)}^2 + \norm{\nabla \ell(\w_i,z_t)}_*^2 } + \frac{R(\wopt)}{\eta (n-1)} 
\intertext{for any non-negative $H$-smooth convex function $f$, we have the self-bounding property that $\norm{\nabla f(\w)}_* \le \sqrt{4 H f(\w)}$. Using this, }
& \le  \frac{ 8 H K^2  \eta }{b^2 (n-1)} \sum_{i=1}^{n-1} \sum_{t=(i-1)b + 1}^{ib}  \E{ L(\w_i) + \ell(\w_i,z_t) } + \frac{R(\wopt)}{\eta (n-1)}  \\
& =  \frac{16 \eta H K^2  }{b } \E{\frac{1}{n-1}\sum_{i=1}^{n-1} L(\w_i) } + \frac{R(\wopt)}{\eta (n-1)} 
\end{align*}
Adding $\frac{1}{n-1}L(\w_1)$ on both sides and removing $L(\w_n)$ on the left we conclude that
\begin{align*}
\E{\frac{1}{n-1} \sum_{i=1}^{n-1} L(\w_{i})}  - L(\wopt) & \le  \frac{16 \eta H K^2 }{b } \E{\frac{1}{n-1}\sum_{i=1}^{n-1} L(\w_i) } + \frac{R(\wopt)}{\eta (n-1)}  + \frac{L(\w_1)}{n-1} 
\end{align*}
Hence we conclude that
\begin{align*}
 \E{\frac{1}{n} \sum_{i=1}^{n} L(\w_{i})}  - L(\wopt) &\le \frac{1}{\left(1 -  \frac{16 \eta H K^2 }{b}\right) } \left( \frac{16 \eta H K^2 }{b} L(\wopt) + \frac{L(\w_1)}{n} + \frac{R(\wopt)}{\eta n}  \right)\\
&= \left( \frac{1}{1 -  \frac{16 \eta H K^2 }{b}} - 1 \right) L(\wopt) + \frac{1}{1 -  \frac{16 \eta H K^2 }{b}} \left(\frac{L(\w_1)}{n} + \frac{R(\wopt)}{\eta n} \right)\\
&= \left( \frac{1}{1 -  \frac{16 \eta H K^2 }{b}} - 1 \right) L(\wopt) + \left(\frac{1}{1 -  \frac{16 \eta H K^2 }{b}}\right) \frac{L(\w_1)}{n} \\
& ~~~~~+ \left(\frac{1}{1 -  \frac{16 \eta H K^2 }{b}}\right)  \frac{b}{16 \eta H K^2 } \frac{16 H K^2  R(\wopt)}{ b n} 
\end{align*}
Writing $\alpha =  \frac{1}{1 -  \frac{16 \eta H K^2 }{b}} - 1 $, so that $\eta = \frac{b}{16 H K^2 }\left(1 - \frac{1}{\alpha + 1}\right)$ we get,
\begin{align*}
 \E{\frac{1}{n} \sum_{i=1}^{n} L(\w_{i})}  - L(\wopt) & \le \alpha L(\wopt) +  \frac{(\alpha + 1) L(\w_1)}{n} + \frac{16 H (\alpha + 1)^2}{\alpha} \frac{R(\wopt)}{b n} \\
 & \le \alpha L(\wopt) +  \frac{(\alpha + 1) L(\w_1)}{n} + \left(\alpha + \frac{1}{\alpha}\right) \frac{32 H R(\wopt)}{b n} 
\end{align*}
Now we shall always pick $\eta \le \frac{b}{32 H K^2 }$ so that $\alpha \le 1$ and so 
\begin{align*}
 \E{\frac{1}{n} \sum_{i=1}^{n} L(\w_{i})}  - L(\wopt) & \le \alpha L(\wopt) + \frac{32 H K^2 R(\wopt)}{\alpha\ b n} +  \frac{2 L(\w_1)}{n} +   \frac{16 H K^2  R(\wopt)}{b n} 
\end{align*}
Picking 
$$
\eta = \min\left\{\frac{1}{2 H}, \frac{b}{32 H K^2 }, \frac{\sqrt{\frac{32 b R(\wopt)}{ L(\wopt) H K^2  n}}}{16\left(1 + \sqrt{\frac{32 H K^2  R(\wopt)}{ L(\wopt) b n}}\right) } \right\} ~ ,
$$
or equivalently  $\alpha = \min\left\{1 , \sqrt{\frac{32 H K^2  R(\wopt)}{ L(\wopt) b n}}\right\}$ we get,
\begin{align*}
 \E{\frac{1}{n} \sum_{i=1}^{n} L(\w_{i})}  - L(\wopt) & \le \sqrt{\frac{128 H K^2  R(\wopt)\ L(\wopt)}{b n}} +  \frac{2 L(\w_1)}{n}  + \frac{16 H K^2  R(\wopt)}{b n} 
\end{align*}
Finally note that by smoothness, 
\begin{align*}
L(\w_1) & \le L(\wopt) + \ip{\nabla L(\w_1) - \nabla L(\wopt)}{\w_1 -\wopt} + \ip{\nabla L(\wopt)}{\w_1 - \wopt} \\
& \le L(\wopt) + \norm{\nabla L(\w_1) - \nabla L(\wopt)}_* \norm{\w_1 -\wopt} + \norm{\nabla L(\wopt)}_* \norm{\w_1 - \wopt} \\
& \le L(\wopt) + H \norm{\w_1 -\wopt}^2 + \sqrt{4 H L(\wopt)} \norm{\w_1 - \wopt} \\
\intertext{Since $R$ is $1$-strongly convex and $\w_1 = \argmin{\w} R(\w)$, }
& \le L(\wopt) + 2 H R(\wopt) + \sqrt{8 H L(\wopt) R(\wopt)} \\
& \le 2 L(\wopt) + 4 H R(\wopt) 
\end{align*}
Hence we conclude that
\begin{align*}
 \E{\frac{1}{n} \sum_{i=1}^{n} L(\w_{i})}  - L(\wopt) & \le \sqrt{\frac{128 H K^2  R(\wopt)\ L(\wopt)}{b n}} +  \frac{4 L(\wopt) + 8 H R(\wopt)}{n} + \frac{16 H K^2  R(\wopt)}{b n}
\end{align*}
Using Jensen's inequality concludes the proof.
\end{proof}

\begin{proof}[Proof of Theorem \ref{thm:sgd}]
For Euclidean case $R(\w) = \frac{1}{2} \norm{\w}_2^2$ and $K = \sqrt{\sup_{\w : \norm{\w}_2 \le 1} \norm{\w}^2} = 1$. 
Plugging these in the previous theorem concludes the proof.
\end{proof}

\subsection{Accelerated Mirror Descent}

\begin{lemma}
For the accelerated update rule, if the step sizes $\beta_i \in [1,\infty)$ and $\gamma_i \in (0,\infty)$ are chosen such that $\beta_1 = 1$ and 
for all $i \in [n]$ 
$$
0 < \gamma_{i+1}(\beta_{i+1} - 1) \le \beta_i \gamma_i ~~~\textrm{and}
~~~ 2 H \gamma_i \le  \beta_i
$$
then we have that
\begin{align*}
\E{L(\w^{\mathrm{ag}}_n)} - L(\wopt) & \le \frac{\gamma_1 (\beta_1 - 1)}{\gamma_n (\beta_n - 1)} L(\w^{\mathrm{ag}}_{1}) + \frac{32 H}{b \gamma_n (\beta_n - 1)} \sum_{i=1}^{n-1} \gamma_i^2 \E{L(\w_i^{\mathrm{ag}})}  + \frac{D^2 }{2 \gamma_n (\beta_n - 1)}  + \frac{16 H^2 D^2}{b \gamma_n (\beta_n - 1)} \sum_{i=1}^{n-1} \frac{\gamma_i^2}{\beta_i^2}
\end{align*}
\end{lemma}
\begin{proof}
First note that for any $i$, 
\begin{align}
\w^{\mathrm{ag}}_{i+1} - \w^{\mathrm{md}}_i & = \beta_i^{-1} \w_{i+1} + (1 - \beta_i^{-1}) \w_{i}^{\mathrm{ag}} - \w^{\mathrm{md}}_i \notag \\
& = \beta_i^{-1} \w_{i+1} + (1 - \beta_i^{-1}) \w_{i}^{\mathrm{ag}} - \beta_i^{-1} \w_{i} - (1 - \beta_i^{-1}) \w_{i}^{\mathrm{ag}} \notag \\
& = \beta_i^{-1} \left(\w_{i+1} - \w_{i}\right) \label{eq:rearrange}
\end{align}
Now by smoothness we have that
\begin{align*}
L(\w^{\mathrm{ag}}_{i+1}) & \le L(\w^{\mathrm{md}}_{i}) + \ip{\nabla L(\w^{\mathrm{md}}_{i})}{\w^{\mathrm{ag}}_{i+1} - \w^{\mathrm{md}}_{i}} + \frac{H}{2}\|\w^{\mathrm{ag}}_{i+1} - \w^{\mathrm{md}}_{i}\|^2 \\
& = L(\w^{\mathrm{md}}_{i}) + \ip{\nabla L(\w^{\mathrm{md}}_{i})}{\w^{\mathrm{ag}}_{i+1} - \w^{\mathrm{md}}_{i}} + \frac{H}{2 \beta_i^2}\|\w_{i+1} - \w_{i}\|^2 \\
& = L(\w^{\mathrm{md}}_{i}) + \ip{\nabla L(\w^{\mathrm{md}}_{i})}{\w^{\mathrm{ag}}_{i+1} - \w^{\mathrm{md}}_{i}} +\frac{1}{2 \beta_i \gamma_i}\|\w_{i+1} - \w_{i}\|^2  - \frac{ \beta_i/\gamma_i - H}{2 \beta_i^2}\|\w_{i+1} - \w_{i}\|^2 \\
\intertext{since $\w_{i+1}^{\mathrm{ag}} = \beta_i^{-1} \w_{i+1} + (1 - \beta_i^{-1}) \w_{i}^{\mathrm{ag}}$, }
& = L(\w^{\mathrm{md}}_{i}) + \ip{\nabla L(\w^{\mathrm{md}}_{i})}{ \beta_i^{-1} \w_{i+1} + (1 - \beta_i^{-1}) \w_{i}^{\mathrm{ag}} - \w^{\mathrm{md}}_{i}} +\frac{\norm{\w_{i} - \w_{i+1}}^2}{ 2 \beta_i \gamma_i}  - \frac{ \beta_i/\gamma_i - H}{2 \beta_i^2}\|\w_{i+1} - \w_{i}\|^2 \\
& = L(\w^{\mathrm{md}}_{i}) + (1 - \beta_i^{-1}) \ip{\nabla L(\w^{\mathrm{md}}_{i})}{\w_{i}^{\mathrm{ag}} - \w^{\mathrm{md}}_{i}}  + \frac{\ip{\nabla L(\w^{\mathrm{md}}_{i})}{ \w_{i+1}  - \w_{i}^{\mathrm{md}}}}{ \beta_i} + \frac{\norm{\w_{i}-\w_{i+1}}^2}{2 \beta_i \gamma_i}  \\
& ~~~~~~~~~~- \frac{ \beta_i/\gamma_i - H}{2 \beta_i^2}\|\w_{i+1} - \w_{i}\|^2 \\
& = (1 - \beta_i^{-1})\left( L(\w^{\mathrm{md}}_{i}) + \ip{\nabla L(\w^{\mathrm{md}}_{i})}{\w_{i}^{\mathrm{ag}} - \w^{\mathrm{md}}_{i}} \right)  + \frac{L(\w^{\mathrm{md}}_i) + \ip{\nabla L(\w^{\mathrm{md}}_{i})}{ \w_{i+1}  - \w_{i}^{\mathrm{md}}}}{\beta_i} + \frac{\norm{\w_{i} - \w_{i+1}}^2}{2 \beta_i \gamma_i}  \\
& ~~~~~~~~~~- \frac{ \beta_i/\gamma_i - H}{2 \beta_i^2}\|\w_{i+1} - \w_{i}\|^2 \\
& = (1 - \beta_i^{-1}) L(\w^{\mathrm{ag}}_{i}) + \frac{L(\w^{\mathrm{md}}_i) + \ip{\nabla L(\w^{\mathrm{md}}_{i})}{ \w_{i+1}  - \w_{i}^{\mathrm{md}}}}{\beta_i} + \frac{\norm{\w_{i}-\w_{i+1}}^2}{2 \beta_i \gamma_i}  - \frac{ \beta_i/\gamma_i - H}{2 \beta_i^2}\|\w_{i+1} - \w_{i}\|^2 \\
\intertext{}
& = (1 - \beta_i^{-1}) L(\w^{\mathrm{ag}}_{i}) - \frac{ \beta_i/\gamma_i - H}{2 \beta_i^2}\|\w_{i+1} - \w_{i}\|^2  + \frac{ \norm{\w_{i} - \w_{i+1}}^2}{2 \beta_i \gamma_i}  \\
& ~~~~~~~~~~ + \frac{L(\w^{\mathrm{md}}_i) + \ip{\nabla \ell_i(\w^{\mathrm{md}}_{i})}{ \w_{i+1}  - \w_{i}^{\mathrm{md}}}  + \ip{\nabla L(\w^{\mathrm{md}}_{i}) - \nabla \ell_i(\w^{\mathrm{md}}_{i})}{ \w_{i+1}  - \w_{i}^{\mathrm{md}}}}{\beta_i} \\
& = (1 - \beta_i^{-1}) L(\w^{\mathrm{ag}}_{i}) - \frac{ \beta_i/\gamma_i - H}{2 \beta_i^2}\|\w_{i+1} - \w_{i}\|^2  + \frac{ \norm{\w_{i} - \w_{i+1}}^2}{2 \beta_i \gamma_i}   + \frac{\ip{\nabla L(\w^{\mathrm{md}}_{i}) - \nabla \ell_i(\w^{\mathrm{md}}_{i})}{ \w_{i+1}  - \w_{i}} }{\beta_i} \\
& ~~~~~~~~~~ + \frac{L(\w^{\mathrm{md}}_i) + \ip{\nabla \ell_i(\w^{\mathrm{md}}_{i})}{ \w_{i+1}  - \w_{i}^{\mathrm{md}}}  + \ip{\nabla L(\w^{\mathrm{md}}_{i}) - \nabla \ell_i(\w^{\mathrm{md}}_{i})}{ \w_{i}  - \w_{i}^{\mathrm{md}}}}{\beta_i}  \\
\intertext{by Holder's inequality,}
& \le (1 - \beta_i^{-1}) L(\w^{\mathrm{ag}}_{i}) - \frac{ \beta_i/\gamma_i - H}{2 \beta_i^2}\|\w_{i+1} - \w_{i}\|^2  + \frac{\norm{\w_{i} - \w_{i+1}}^2}{ 2 \beta_i \gamma_i}  + \frac{ \norm{\nabla L(\w^{\mathrm{md}}_{i}) - \nabla \ell_i(\w^{\mathrm{md}}_{i})}_* \norm{ \w_{i+1}  - \w_{i}} }{\beta_i}\\
& ~~~~~~~~~~ + \frac{L(\w^{\mathrm{md}}_i) + \ip{\nabla \ell_i(\w^{\mathrm{md}}_{i})}{ \w_{i+1}  - \w_{i}^{\mathrm{md}}}  + \ip{\nabla L(\w^{\mathrm{md}}_{i}) - \nabla \ell_i(\w^{\mathrm{md}}_{i})}{ \w_{i}  - \w_{i}^{\mathrm{md}}}}{\beta_i} \\
\intertext{since for any $a,b$ and $\alpha > 0$, $ab \le \frac{a^2}{2 \alpha} + \frac{\alpha b^2}{2}$}
& \le (1 - \beta_i^{-1}) L(\w^{\mathrm{ag}}_{i}) + \frac{\norm{\nabla L(\w^{\mathrm{md}}_{i}) - \nabla \ell_i(\w^{\mathrm{md}}_{i})}_*^2}{2 ( \beta_i/\gamma_i - H)} + \frac{\norm{\w_{i} - \w_{i+1}}^2}{2 \beta_i \gamma_i}  \\
& ~~~~~~~~~~ + \frac{L(\w^{\mathrm{md}}_i) + \ip{\nabla \ell_i(\w^{\mathrm{md}}_{i})}{ \w_{i+1}  - \w_{i}^{\mathrm{md}}}  + \ip{\nabla L(\w^{\mathrm{md}}_{i}) - \nabla \ell_i(\w^{\mathrm{md}}_{i})}{ \w_{i}  - \w_{i}^{\mathrm{md}}}}{\beta_i} 
\end{align*}
We now note that the update step 2 of accelerated gradient can be written equivalently as 
$$\w_{i+1} = \argmin{\w \in \Wcal} \left\{\gamma_i \ip{\nabla \ell_i(\w_i^{\mathrm{md}})}{\w - \w_i^{\mathrm{md}}} + \breg{R}{\w}{\w_{i}^{\mathrm{md}}}\right\}~.$$ 
It can be shown that (see for instance Lemma 1 of \cite{Lan09}) 
$$
\gamma_i  \ip{\nabla \ell_i(\w_i)}{\w_{i+1} - \w_i^{\mathrm{md}}} \le \gamma_i  \ip{\nabla \ell_i(\w_i^{\mathrm{md}})}{\wopt - \w_i^{\mathrm{md}}} + \breg{R}{\wopt}{\w_{i}} - \breg{R}{\wopt}{\w_{i+1}} - \breg{R}{\w_i}{\w_{i+1}}
$$
Plugging this we get that,
\begin{align*}
L(\w_{i+1}^{\mathrm{ag}}) & \le (1 - \beta_i^{-1}) L(\w^{\mathrm{ag}}_{i}) + \frac{\norm{\nabla L(\w^{\mathrm{md}}_{i}) - \nabla \ell_i(\w^{\mathrm{md}}_{i})}_*^2}{2 ( \beta_i/\gamma_i - H)} + \frac{\norm{\w_{i} - \w_{i+1}}^2}{2 \beta_i \gamma_i}  + \frac{ \ip{\nabla \ell_i(\w^{\mathrm{md}}_{i})}{ \wopt  - \w_{i}^{\mathrm{md}}} }{\beta_i} \\
& ~~~~~~~~ + \frac{L(\w^{\mathrm{md}}_i)  + \ip{\nabla L(\w^{\mathrm{md}}_{i}) - \nabla \ell_i(\w^{\mathrm{md}}_{i})}{ \w_{i}  - \w_{i}^{\mathrm{md}}}}{\beta_i} + \frac{\breg{R}{\wopt}{\w_{i}} - \breg{R}{\wopt}{\w_{i+1}} - \breg{R}{\w_i}{\w_{i+1}}}{\gamma_i \beta_i}
\intertext{by strong-convexity of $R$, $\breg{R}{\w_i}{\w_{i+1}} \ge \frac{1}{2}\norm{\w_i- \w_{i+1}}^2$ and so,}
& = (1 - \beta_i^{-1}) L(\w^{\mathrm{ag}}_{i}) + \frac{\norm{\nabla L(\w^{\mathrm{md}}_{i}) - \nabla \ell_i(\w^{\mathrm{md}}_{i})}_*^2}{2 ( \beta_i/\gamma_i - H)}  + \frac{ \ip{\nabla \ell_i(\w^{\mathrm{md}}_{i})}{ \wopt  - \w_{i}^{\mathrm{md}}} }{\beta_i} \\
& ~~~~~~~~ + \frac{L(\w^{\mathrm{md}}_i)  + \ip{\nabla L(\w^{\mathrm{md}}_{i}) - \nabla \ell_i(\w^{\mathrm{md}}_{i})}{ \w_{i}  - \w_{i}^{\mathrm{md}}}}{\beta_i} + \frac{\breg{R}{\wopt}{\w_{i}} - \breg{R}{\wopt}{\w_{i+1}} }{\gamma_i \beta_i} \\
& = (1 - \beta_i^{-1}) L(\w^{\mathrm{ag}}_{i}) + \frac{\norm{\nabla L(\w^{\mathrm{md}}_{i}) - \nabla \ell_i(\w^{\mathrm{md}}_{i})}_*^2}{2 ( \beta_i/\gamma_i - H)}  + \frac{ \ip{\nabla \ell_i(\w^{\mathrm{md}}_{i}) - \nabla L(\w_i^{\mathrm{md}})}{ \wopt  - \w_{i}^{\mathrm{md}}} }{\beta_i} \\
& ~~~~~~~~ + \frac{\ip{\nabla L(\w^{\mathrm{md}}_{i}) - \nabla \ell_i(\w^{\mathrm{md}}_{i})}{ \w_{i}  - \w_{i}^{\mathrm{md}}}}{\beta_i} + \frac{\breg{R}{\wopt}{\w_{i}} - \breg{R}{\wopt}{\w_{i+1}} }{\gamma_i \beta_i} \\
& ~~~~~~~~ + \frac{ L(\w^{\mathrm{md}}_i) + \ip{\nabla L(\w_i^{\mathrm{md}})}{ \wopt  - \w_{i}^{\mathrm{md}}} }{\beta_i}  \\
\intertext{by convexity, $L(\wopt) \ge L(\w^{\mathrm{md}}_i) + \ip{\nabla L(\w_i^{\mathrm{md}})}{\wopt - \w^{\mathrm{md}}_i}$, hence }
& \le (1 - \beta_i^{-1}) L(\w^{\mathrm{ag}}_{i}) + \frac{\norm{\nabla L(\w^{\mathrm{md}}_{i}) - \nabla \ell_i(\w^{\mathrm{md}}_{i})}_*^2}{2 ( \beta_i/\gamma_i - H)}  + \frac{ \ip{\nabla \ell_i(\w^{\mathrm{md}}_{i}) - \nabla L(\w_i^{\mathrm{md}})}{ \wopt  - \w_{i}^{\mathrm{md}}} }{\beta_i} \\
& ~~~~~~~~ + \frac{\ip{\nabla L(\w^{\mathrm{md}}_{i}) - \nabla \ell_i(\w^{\mathrm{md}}_{i})}{ \w_{i}  - \w_{i}^{\mathrm{md}}}}{\beta_i} + \frac{\breg{R}{\wopt}{\w_{i}} - \breg{R}{\wopt}{\w_{i+1}} }{ \gamma_i \beta_i}  + \frac{ L(\wopt) }{\beta_i}  \\
& = (1 - \beta_i^{-1}) L(\w^{\mathrm{ag}}_{i}) + \frac{\norm{\nabla L(\w^{\mathrm{md}}_{i}) - \nabla \ell_i(\w^{\mathrm{md}}_{i})}_*^2}{2 ( \beta_i/\gamma_i - H)} + \frac{\ip{\nabla L(\w^{\mathrm{md}}_{i}) - \nabla \ell_i(\w^{\mathrm{md}}_{i})}{ \w_{i}  - \wopt}}{\beta_i} \\  
& ~~~~~~~~  + \frac{\breg{R}{\wopt}{\w_{i}} - \breg{R}{\wopt}{\w_{i+1}} }{\gamma_i \beta_i}  + \beta_i^{-1} L(\wopt)   \\
& = L(\wopt) +  (1 - \beta_i^{-1}) \left(L(\w^{\mathrm{ag}}_{i}) - L(\wopt) \right) + \frac{\norm{\nabla L(\w^{\mathrm{md}}_{i}) - \nabla \ell_i(\w^{\mathrm{md}}_{i})}_*^2}{2 ( \beta_i/\gamma_i - H)} + \frac{\ip{\nabla L(\w^{\mathrm{md}}_{i}) - \nabla \ell_i(\w^{\mathrm{md}}_{i})}{ \w_{i}  - \wopt}}{\beta_i} \\  
& ~~~~~~~~  + \frac{\breg{R}{\wopt}{\w_{i}} - \breg{R}{\wopt}{\w_{i+1}} }{\gamma_i \beta_i}
\end{align*}
Thus we conclude that
\begin{align*}
L(\w^{\mathrm{ag}}_{i+1}) - L(\wopt) & \le  (1 - \beta_i^{-1})\left( L(\w^{\mathrm{ag}}_{i}) - L(\wopt)\right)+ \frac{\norm{\nabla L(\w^{\mathrm{md}}_{i}) - \nabla \ell_i(\w^{\mathrm{md}}_{i})}_*^2}{2 ( \beta_i/\gamma_i - H)} + \frac{\ip{\nabla L(\w^{\mathrm{md}}_{i}) - \nabla \ell_i(\w^{\mathrm{md}}_{i})}{ \w_{i}  - \wopt}}{\beta_i}  \\
& ~~~~~~  + \frac{\breg{R}{\wopt}{\w_{i}} - \breg{R}{\wopt}{\w_{i+1}} }{ \beta_i \gamma_i} 
\end{align*}
Multiplying throughout by $\beta_i \gamma_i$ we get
\begin{align*}
\gamma_i \beta_i \left(L(\w^{\mathrm{ag}}_{i+1}) - L(\wopt) \right) & \le   \gamma_i (\beta_i - 1)\left( L(\w^{\mathrm{ag}}_{i}) - L(\wopt)\right)+ \frac{\gamma_i \beta_i \norm{\nabla L(\w^{\mathrm{md}}_{i}) - \nabla \ell_i(\w^{\mathrm{md}}_{i})}_*^2}{2 ( \beta_i/\gamma_i - H)}  \\
& ~~~~~~  + \breg{R}{\wopt}{\w_{i}} - \breg{R}{\wopt}{\w_{i+1}}  + \gamma_i \ip{\nabla L(\w^{\mathrm{md}}_{i}) - \nabla \ell_i(\w^{\mathrm{md}}_{i})}{ \w_{i}  - \wopt}
\end{align*}

Owing to the condition that $\gamma_{i+1} (\beta_{i+1} -1) \le \gamma_i \beta_i$ we have that 
\begin{align*}
\gamma_{i+1}(\beta_{i+1} -1) \left(L(\w^{\mathrm{ag}}_{i+1}) - L(\wopt) \right) & \le   \gamma_i (\beta_i - 1)\left( L(\w^{\mathrm{ag}}_{i}) - L(\wopt)\right)+ \frac{\gamma_i \beta_i \norm{\nabla L(\w^{\mathrm{md}}_{i}) - \nabla \ell_i(\w^{\mathrm{md}}_{i})}_*^2}{2 ( \beta_i/\gamma_i - H)}  \\
& ~~~~~~  + \breg{R}{\wopt}{\w_{i}} - \breg{R}{\wopt}{\w_{i+1}}   + \gamma_i \ip{\nabla L(\w^{\mathrm{md}}_{i}) - \nabla \ell_i(\w^{\mathrm{md}}_{i})}{ \w_{i}  - \wopt}
\end{align*}
Using the above inequality repeatedly we conclude that
\begin{align*}
\gamma_{n}(\beta_{n} - 1) \left(L(\w^{\mathrm{ag}}_{n}) - L(\wopt) \right) & \le   \gamma_1 (\beta_1 - 1)\left( L(\w^{\mathrm{ag}}_{1}) - L(\wopt)\right)+ \sum_{i=1}^{n-1} \frac{\gamma_i \beta_i \norm{\nabla L(\w^{\mathrm{md}}_{i}) - \nabla \ell_i(\w^{\mathrm{md}}_{i})}_*^2}{2 ( \beta_i/\gamma_i - H)}  \\
& ~~~~~~  + \breg{R}{\wopt}{\w_{1}} - \breg{R}{\wopt}{\w_{n}}  + \sum_{i=1}^{n-1} \gamma_i \ip{\nabla L(\w^{\mathrm{md}}_{i}) - \nabla \ell_i(\w^{\mathrm{md}}_{i})}{ \w_{i}  - \wopt}\\
& \le   \gamma_1 (\beta_1 - 1)\left( L(\w^{\mathrm{ag}}_{1}) - L(\wopt)\right)+ \sum_{i=1}^{n-1} \frac{\gamma_i \beta_i \norm{\nabla L(\w^{\mathrm{md}}_{i}) - \nabla \ell_i(\w^{\mathrm{md}}_{i})}_*^2}{2 ( \beta_i/\gamma_i - H)}  \\
& ~~~~~~  + R(\wopt) + \sum_{i=1}^{n-1} \gamma_i \ip{\nabla L(\w^{\mathrm{md}}_{i}) - \nabla \ell_i(\w^{\mathrm{md}}_{i})}{ \w_{i}  - \wopt}\\
& =   \gamma_1 (\beta_1 - 1)\left( L(\w^{\mathrm{ag}}_{1}) - L(\wopt)\right)+ \sum_{i=1}^{n-1} \frac{\gamma_i \beta_i \norm{\nabla L(\w^{\mathrm{md}}_{i}) - \nabla \ell_i(\w^{\mathrm{md}}_{i})}_*^2}{2 ( \beta_i/\gamma_i - H)}  \\
& ~~~~~~  + R(\wopt) + \sum_{i=1}^{n-1} \gamma_i \ip{\nabla L(\w^{\mathrm{md}}_{i}) - \nabla \ell_i(\w^{\mathrm{md}}_{i})}{ \w_{i}  - \wopt}
\intertext{since $2 H \gamma_i \le \beta_i$, }
& \le   \gamma_1 (\beta_1 - 1)\left( L(\w^{\mathrm{ag}}_{1}) - L(\wopt)\right)+ \sum_{i=1}^{n-1} \gamma_i^2 \norm{\nabla L(\w^{\mathrm{md}}_{i}) - \nabla \ell_i(\w^{\mathrm{md}}_{i})}_*^2 + R(\wopt) \\
& ~~~~~~   + \sum_{i=1}^{n-1} \gamma_i \ip{\nabla L(\w^{\mathrm{md}}_{i}) - \nabla \ell_i(\w^{\mathrm{md}}_{i})}{ \w_{i}  - \wopt}\\
& \le   \gamma_1 (\beta_1 - 1) L(\w^{\mathrm{ag}}_{1}) + \sum_{i=1}^{n-1} 2 \gamma_i^2 \norm{\nabla L(\w^{\mathrm{ag}}_{i}) - \nabla \ell_i(\w^{\mathrm{ag}}_{i})}_*^2 + R(\wopt) \\
& ~~~~~~   + \sum_{i=1}^{n-1} \gamma_i \ip{\nabla L(\w^{\mathrm{md}}_{i}) - \nabla \ell_i(\w^{\mathrm{md}}_{i})}{ \w_{i}  - \wopt} \\
& ~~~~~~ + \sum_{i=1}^{n-1} 2 \gamma_i^2 \norm{\nabla L(\w^{\mathrm{md}}_{i}) - \nabla \ell_i(\w^{\mathrm{md}}_{i}) - \nabla L(\w^{\mathrm{ag}}_{i}) + \nabla \ell_i(\w^{\mathrm{ag}}_{i})}_*^2
\end{align*}
Taking expectation we get that
\begin{align}
\gamma_{n}(\beta_{n} - 1) \left(\E{L(\w^{\mathrm{ag}}_{n})} - L(\wopt) \right) & \le   \gamma_1 (\beta_1 - 1) L(\w^{\mathrm{ag}}_{1}) + \sum_{i=1}^{n-1} 2 \gamma_i^2 \E{\norm{\nabla L(\w^{\mathrm{ag}}_{i}) - \nabla \ell_i(\w^{\mathrm{ag}}_{i})}_*^2} + R(\wopt) \notag \\
& ~~~~~~ + \sum_{i=1}^{n-1} 2 \gamma_i^2 \E{\norm{\nabla L(\w^{\mathrm{md}}_{i}) - \nabla \ell_i(\w^{\mathrm{md}}_{i}) - \nabla L(\w^{\mathrm{ag}}_{i}) + \nabla \ell_i(\w^{\mathrm{ag}}_{i})}_*^2} \label{eq:inter}
\end{align}
Now note that 
$$
\nabla L(\w^{\mathrm{ag}}_i) - \nabla \ell_i(\w^{\mathrm{ag}}_i) = \frac{1}{b} \sum_{t= (i-1)b+ 1}^{bi} \left(\nabla L(\w^{\mathrm{ag}}_i) - \ell(\w^{\mathrm{ag}}_i,z_t)\right) ~~~~~\textrm{and}
$$
$$
\nabla L(\w^{\mathrm{ag}}_i) - \nabla \ell_i(\w^{\mathrm{ag}}_i) - \nabla L(\w^{\mathrm{md}}_i) + \nabla \ell_i(\w^{\mathrm{md}}_i) = \frac{1}{b} \sum_{t= (i-1)b+ 1}^{bi} \left(\nabla L(\w^{\mathrm{ag}}_i) - \ell(\w^{\mathrm{ag}}_i,z_t) - \nabla L(\w^{\mathrm{md}}_i) + \ell(\w^{\mathrm{md}}_i,z_t)\right)
$$
Further $\left(\nabla L(\w_i) - \ell(\w_i,z_t)\right)$ and $\left(\nabla L(\w^{\mathrm{ag}}_i) - \ell(\w^{\mathrm{ag}}_i,z_t) - \nabla L(\w^{\mathrm{md}}_i) + \ell(\w^{\mathrm{md}}_i,z_t)\right)$ are mean zero vectors drawn i.i.d. Also note that $\w^{\mathrm{ag}}_i$ only depends on the first $(i-1)b$ examples and so when we consider expectation w.r.t. $z_{(i-1)b + 1} , \ldots, z_{ib}$, $\w_i$ is fixed.  Hence by Corollary \ref{cor:eqsmooth} we have that,
\begin{align*}
\E{\norm{\nabla L(\w_i^{\mathrm{ag}}) - \nabla \ell_i(\w_i^{\mathrm{ag}})}_*^2} & = \frac{K^2}{b^2} \E{\norm{\sum_{t= (i-1)b+ 1}^{bi} \left(\nabla L(\w_i^{\mathrm{ag}}) - \nabla  \ell(\w_i^{\mathrm{ag}},z_t)\right)}_*^2}\\
& \le \frac{K^2}{b^2} \sum_{t= (i-1)b+ 1}^{bi} \E{\norm{ \left(\nabla L(\w_i^{\mathrm{ag}}) - \nabla \ell(\w_i^{\mathrm{ag}},z_t)\right)}_*^2}
\end{align*}
and similarly 
\begin{align*}
& \E{\norm{\nabla L(\w^{\mathrm{ag}}_i) - \nabla \ell_i(\w^{\mathrm{ag}}_i) - \nabla L(\w^{\mathrm{md}}_i) + \nabla \ell_i(\w^{\mathrm{md}}_i)}_*^2} \\
& ~~~~~~~~~~ \le  \frac{K^2}{b^2} \sum_{t= (i+1)b +1}^{bi} \E{\norm{\nabla L(\w^{\mathrm{ag}}_i) - \nabla \ell(\w^{\mathrm{ag}}_i,z_t) - \nabla L(\w^{\mathrm{md}}_i) + \nabla \ell(\w^{\mathrm{md}}_i, z_t)}_*^2}
\end{align*}
Plugging these back in Equation \ref{eq:inter} we get :
{\small
\begin{align*}
\gamma_{n}(\beta_{n} - 1) \left(\E{L(\w^{\mathrm{ag}}_{n})} - L(\wopt) \right) & \le   \gamma_1 (\beta_1 - 1) L(\w^{\mathrm{ag}}_{1}) + \sum_{i=1}^{n-1}  \frac{2 K^2 \gamma_i^2}{b^2} \sum_{t= (i-1)b+ 1}^{bi} \E{\norm{ \left(\nabla L(\w_i^{\mathrm{ag}}) - \nabla \ell(\w_i^{\mathrm{ag}},z_t)\right)}_*^2}  + R(\wopt)  \\
& ~~~ + \sum_{i=1}^{n-1} \frac{2 K^2 \gamma_i^2}{b^2} \sum_{t= (i+1)b +1}^{bi} \E{\norm{\nabla L(\w^{\mathrm{ag}}_i) - \nabla \ell(\w^{\mathrm{ag}}_i,z_t) - \nabla L(\w^{\mathrm{md}}_i) + \nabla \ell(\w^{\mathrm{md}}_i, z_t)}_*^2}\\
& \le   \gamma_1 (\beta_1 - 1) L(\w^{\mathrm{ag}}_{1}) + \sum_{i=1}^{n-1}  \frac{4 K^2 \gamma_i^2}{b^2} \sum_{t= (i-1)b+ 1}^{bi} \E{\norm{ \nabla L(\w_i^{\mathrm{ag}})}_*^2 + \norm{\nabla \ell(\w_i^{\mathrm{ag}},z_t)}_*^2}  + R(\wopt) \\
& ~~~ + \sum_{i=1}^{n-1} \frac{4 K^2  \gamma_i^2}{b^2} \sum_{t= (i+1)b +1}^{bi} \E{\norm{\nabla L(\w^{\mathrm{ag}}_i)  - \nabla L(\w^{\mathrm{md}}_i)}_*^2 + \norm{\nabla \ell(\w^{\mathrm{md}}_i, z_t) - \nabla \ell(\w^{\mathrm{ag}}_i,z_t)}_*^2}
\intertext{for any non-negative $H$-smooth convex function $f$, we have the self-bounding property that $\norm{\nabla f(\w)} \le \sqrt{4 H f(\w)}$. Using this, }
& \le   \gamma_1 (\beta_1 - 1) L(\w^{\mathrm{ag}}_{1}) + \sum_{i=1}^{n-1}  \frac{16 H K^2  \gamma_i^2}{b^2} \sum_{t= (i-1)b+ 1}^{bi} \E{L(\w_i^{\mathrm{ag}}) + \ell(\w_i^{\mathrm{ag}},z_t)}  + R(\wopt) \\
& ~~~ + \sum_{i=1}^{n-1} \frac{4 K^2 \gamma_i^2}{b^2 } \sum_{t= (i+1)b +1}^{bi} \E{\norm{\nabla L(\w^{\mathrm{ag}}_i)  - \nabla L(\w^{\mathrm{md}}_i)}_*^2 + \norm{\nabla \ell(\w^{\mathrm{md}}_i, z_t) - \nabla \ell(\w^{\mathrm{ag}}_i,z_t)}_*^2}\\
& =   \gamma_1 (\beta_1 - 1) L(\w^{\mathrm{ag}}_{1}) + \sum_{i=1}^{n-1}  \frac{32 H K^2  \gamma_i^2}{b} \E{L(\w_i^{\mathrm{ag}})}  + R(\wopt)\\
& ~~~ + \sum_{i=1}^{n-1} \frac{4 K^2  \gamma_i^2}{b^2 } \sum_{t= (i+1)b +1}^{bi} \E{\norm{\nabla L(\w^{\mathrm{ag}}_i)  - \nabla L(\w^{\mathrm{md}}_i)}_*^2 + \norm{\nabla \ell(\w^{\mathrm{md}}_i, z_t) - \nabla \ell(\w^{\mathrm{ag}}_i,z_t)}_*^2}
\intertext{by $H$-smoothness of $L$ and $\ell$ we have that $\norm{\nabla L(\w^{\mathrm{ag}}_i)  - \nabla L(\w^{\mathrm{md}}_i)}_* \le H \norm{\w^{\mathrm{ag}}_i - \w^{\mathrm{md}}_i}$. Similarly we also have that $\norm{\nabla \ell(\w^{\mathrm{ag}}_i,z_t)  - \nabla \ell(\w^{\mathrm{md}}_i,z_t)}_* \le H \norm{\w^{\mathrm{ag}}_i - \w^{\mathrm{md}}_i}$. Hence,}
& \le \gamma_1 (\beta_1 - 1) L(\w^{\mathrm{ag}}_{1}) + \sum_{i=1}^{n-1}  \frac{32 H K^2 \gamma_i^2}{b} \E{L(\w_i^{\mathrm{ag}})}  + R(\wopt) \\
& ~~~ + \sum_{i=1}^{n-1} \frac{8 H^2 K^2  \gamma_i^2}{b} \E{\norm{\w^{\mathrm{ag}}_i - \w^{\mathrm{md}}_i}^2 }
\intertext{However, $\w^{\mathrm{md}}_i \leftarrow \beta_i^{-1} \w_{i} + (1 - \beta_i^{-1}) \w_{i}^{\mathrm{ag}}$. Hence $\norm{\w^{\mathrm{ag}}_i - \w^{\mathrm{md}}_i}^2 \le \frac{\norm{\w_i - \w_{i}^{\mathrm{ag}}}^2}{\beta_i^2} \le \frac{2 \norm{\w_i - \w_1}^2 + 2 \norm{\w_1- \w_{i}^{\mathrm{ag}}}^2}{\beta_i^2} \le \frac{4 D^2}{\beta_i^2}$ . Hence,}
& \le \gamma_1 (\beta_1 - 1) L(\w^{\mathrm{ag}}_{1}) + \sum_{i=1}^{n-1}  \frac{32 H K^2 \gamma_i^2}{b} \E{L(\w_i^{\mathrm{ag}})}  + R(\wopt)  + \frac{32 H^2 K^2  D^2}{b} \sum_{i=1}^{n-1} \frac{\gamma_i^2}{\beta_i^2}
\end{align*}
}
Dividing throughout by $\gamma_n(\beta_n -1)$ concludes the proof.

\end{proof}

\begin{proof}[Proof of Theorem \ref{thm:amd}]
First note that the for any $i$,
$$
2 H \gamma_i = 2 H \gamma i^{p} \le \frac{i^p}{2} \le \beta_i 
$$
Also note that since $p \in [0,1]$,
$$
\gamma_{i+1} (\beta_{i+1} -1) = \gamma \frac{i (i+1)^p}{2} \le \gamma \frac{i^p (i+1)}{2} = \gamma_i \beta_i
$$
Thus we have verified that the step sizes satisfy the conditions required by previous lemma. From the previous lemma we have that
\begin{align*}
\E{L(\w^{\mathrm{ag}}_n)} - L(\wopt) & \le  \frac{\gamma_1 (\beta_1 - 1)}{\gamma_n (\beta_n - 1)} L(\w^{\mathrm{ag}}_{1}) + \frac{32 H K^2 }{b \gamma_n (\beta_n - 1)} \sum_{i=1}^{n-1} \gamma_i^2 \E{L(\w_i^{\mathrm{ag}})}  + \frac{R(\wopt)}{ \gamma_n (\beta_n - 1)}  + \frac{32 H^2 K^2  D^2}{b \gamma_n (\beta_n - 1)} \sum_{i=1}^{n-1} \frac{\gamma_i^2}{\beta_i^2}\\
& = \frac{64 H K^2 \gamma}{b n^p (n - 1)} \sum_{i=1}^{n-1} i^{2p}\ \E{L(\w_i^{\mathrm{ag}})}  + \frac{2 R(\wopt) }{\gamma n^p (n - 1)}  + \frac{256 H^2 K^2 D^2 \gamma}{b n^p (n - 1)} \sum_{i=1}^{n-1} \frac{i^{2p}}{(i+1)^2}\\
& \le \frac{64 H K^2 \gamma (n-1)^{2p}}{b n^p (n - 1)} \sum_{i=1}^{n-1}  \E{L(\w_i^{\mathrm{ag}})}  + \frac{2 R(\wopt)}{\gamma  (n - 1)^{p+1}}  + \frac{256 H^2 K^2 D^2 \gamma}{b  (n - 1)^{p+1}} \sum_{i=1}^{n-1} \frac{1}{i^{2(1 - p)}}\\
& \le \frac{64 H K^2 \gamma }{b  (n - 1)^{1 - p}} \sum_{i=1}^{n-1}  \E{L(\w_i^{\mathrm{ag}})}  + \frac{2 R(\wopt)}{\gamma (n - 1)^{p+1}}  + \frac{256 H^2 K^2 D^2 \gamma}{b  (n - 1)^{p+1}} \sum_{i=1}^{n-1} \frac{1}{i^{2(1-p)}}\\
& \le \frac{64 H K^2 \gamma }{b  (n - 1)^{1 - p}} \sum_{i=1}^{n-1}  \E{L(\w_i^{\mathrm{ag}})}  + \frac{2 R(\wopt)}{\gamma (n - 1)^{p+1}}  + \frac{256 H^2 K^2 D^2 \gamma}{b  (n - 1)}\\
& \le \frac{64 H K^2 \gamma }{b  (n - 1)^{1 - p}} \sum_{i=1}^{n-1}  \left(\E{L(\w_i^{\mathrm{ag}})} - L(\wopt)\right) +  \frac{64 H K^2 \gamma L(\wopt) (n - 1)^{p}}{b}  + \frac{2 R(\wopt)}{\gamma (n - 1)^{p+1}}  \\
& ~~~~~+ \frac{256 H^2 K^2 D^2 \gamma}{b  (n - 1)}
\intertext{since $\gamma \le 1/4H$,}
& \le \frac{64 H K^2 \gamma }{b  (n - 1)^{1 - p}} \sum_{i=1}^{n-1}  \left(\E{L(\w_i^{\mathrm{ag}})} - L(\wopt)\right) +  \frac{64 H K^2 \gamma L(\wopt) (n - 1)^{p}}{b}  + \frac{2 R(\wopt)}{\gamma (n - 1)^{p+1}}\\
& ~~~~~  + \frac{64 H K^2 D^2}{b  (n - 1)}
\end{align*}
Thus we have shown that
\begin{align*}
\E{L(\w^{\mathrm{ag}}_n)} - L(\wopt) & \le \frac{64 H K^2 \gamma }{b  (n - 1)^{1 - p}} \sum_{i=1}^{n-1}  \left(\E{L(\w_i^{\mathrm{ag}})} - L(\wopt)\right) +  \frac{64 H K^2 \gamma L(\wopt) (n - 1)^{p}}{b}  + \frac{2 R(\wopt)}{\gamma (n - 1)^{p+1}}  \\
& ~~~~~ + \frac{64 H K^2 D^2}{b  (n - 1)}
\end{align*}

Now if we use the notation $a_i = \E{L(\w_i^{\mathrm{ag}})} - L(\wopt)$,
$A(i) = \frac{64 H K^2 \gamma }{b  (i - 1)^{1 - p}}$ and 
$$
B(i) = \frac{64 H K^2 \gamma L(\wopt) (i - 1)^{p}}{b}  + \frac{2 R(\wopt)}{\gamma (i - 1)^{p+1}}  + \frac{64 H K^2 D^2}{b  (i - 1)}
$$ 
Note that for any $i$ by smoothness, $a_i  \le  L_0 := \frac{3}{2} H D^2 + L(\wopt)$
Also notice that 
\begin{align*}
\sum_{i=n - M - 1}^n A(i)  & = \frac{64 H K^2 \gamma }{b} \sum_{i=n-M-1}^n \frac{1}{(i - 1)^{1 - p}}  \le \frac{64 H K^2 \gamma n^p}{b} 
\end{align*}
Hence as long as 
\begin{align}\label{eq:con1}
\gamma \le \frac{b}{64 H K^2 n^p}~,
\end{align}
$\sum_{i=n - M - 1}^n A(i) \le 1$. We shall ensure that the $\gamma$ we choose will satisfy the above condition. Now applying lemma \ref{ut:rec} we get that for any $M$,
\begin{align}\label{eq:fromrec}
a_n & \le e A(n)\left( a_0 (n-M) + \sum_{i= n-M-1}^n B(i) \right) + B(n)
\end{align}
Now notice that 
\begin{align*}
\sum_{i= n-M-1}^n B(i)  &= \frac{64 H K^2 \gamma L(\wopt) }{b} \sum_{i= n-M-1}^n \frac{1}{(i-1)^p}  + \frac{2 R(\wopt)}{\gamma} \sum_{i=n-M-1}^n \frac{1}{(i - 1)^{p+1}}  + \frac{64 H K^2 D^2}{b}  \sum_{i=n-M-1}^n \frac{1}{(i-1)}\\
& \le \frac{64 H K^2 \gamma L(\wopt) (n-M-2)^{p}}{b}  + \frac{2 R(\wopt)}{\gamma (n-M-2)^{p+1}}  + \frac{64 H K^2 D^2}{b  (n-M-2)} + \frac{64 H K^2 \gamma L(\wopt) (n-1)^{p+1}}{b}\\
&~~~~~~~~~~  + \frac{2 R(\wopt)}{\gamma (n -M - 2)^{p}}  + \frac{64 H K^2 D^2 \log\ n}{b }
\end{align*}
Plugging this back in Equation \ref{eq:fromrec} we conclude that
\begin{align*}
a_n & \le \frac{64 e H K^2 \gamma }{b  (n - 1)^{1 - p}}\Big( L_0 (n-M)  + \frac{64 H K^2 \gamma L(\wopt) (n-M-2)^{p}}{b}  + \frac{2 R(\wopt)}{\gamma (n-M-2)^{p+1}}  \\
& ~~~~~ + \frac{64 H K^2 D^2}{b  (n-M-2)} + \frac{64 H K^2 \gamma L(\wopt) (n-1)^{p+1}}{b} + \frac{2 R(\wopt)}{\gamma (n -M - 2)^{p}}  + \frac{64 H K^2 D^2 \log\ n}{b} \Big) \\
& ~~~~~~~~~+ \frac{64 H K^2 \gamma L(\wopt) (n - 1)^{p}}{b}  + \frac{2 R(\wopt)}{\gamma (n - 1)^{p+1}}  + \frac{64 H K^2 D^2}{b  (n - 1)}\\
& \le \frac{64 e H K^2 \gamma }{b  (n - 1)^{1 - p}}\Big( L_0 (n-M - 2)   +  \frac{64 H K^2 D^2}{b  (n-M-2)} +  \frac{4 R(\wopt)}{\gamma (n -M - 2)^{p}}  + \frac{64 H K^2 D^2 \log(n)}{b}\\
& ~~~~~ + \frac{256 H K^2 \gamma L(\wopt) (n-1)^{p+1}}{b}   \Big) + \frac{64 H K^2  \gamma L(\wopt) (n - 1)^{p}}{b}  + \frac{4 R(\wopt)}{\gamma (n - 1)^{p+1}}  + \frac{64 H K^2 D^2}{b  (n - 1)}
\intertext{since $\gamma \le \frac{b}{64 H K^2 n^p}$ and $ \frac{64 H K^2 D^2}{b  (n-M-2)} \le  \frac{4 R(\wopt) }{\gamma (n -M - 2)^{p}}$, }
& \le \frac{64 e H K^2 \gamma }{b  (n - 1)^{1 - p}}\Big( L_0 (n-M - 2)   +  \frac{6 R(\wopt) }{\gamma (n -M - 2)^{p}} + \frac{64 H K^2 D^2 \log\ n}{b}  \\
& ~~~~~ + \frac{256 H K^2 \gamma L(\wopt) (n-1)^{p+1}}{b}   \Big) + \frac{64 H K^2 \gamma L(\wopt) (n - 1)^{p}}{b}  + \frac{4 R(\wopt)}{\gamma (n - 1)^{p+1}}  + \frac{64 H K^2 D^2}{b  (n - 1)}
\end{align*}
We now optimize over the choice of $M$ above by using 
$$
(n - M - 2) = \left(\frac{6 R(\wopt)}{\gamma L_0}\right)^{\frac{1}{p+1}}
$$
Ofcourse for the choice of $M$ to be valid we need that 
$n - M - 2 \le n$ which gives our second condition on $\gamma$ which is
\begin{align}\label{eq:con2}
\gamma \ge \frac{6 R(\wopt)}{n^{p+1} L_0}
\end{align}
Plugging in this $M$ we get,
\begin{align}
a_n & \le \frac{64 e H K^2 \gamma }{b  (n - 1)^{1 - p}}\left( 2 L_0^{\frac{p}{p+1}} \left(\frac{6 R(\wopt)}{\gamma }\right)^{\frac{1}{p+1}}  + \frac{128 H K^2 \gamma L(\wopt) (n-1)^{p+1}}{b}   + \frac{64 H K^2 D^2 \log\ n}{b} \right) \notag \\
& ~~~~~+ \frac{64 H K^2 \gamma L(\wopt) (n - 1)^{p}}{b}  + \frac{2 R(\wopt)}{\gamma (n - 1)^{p+1}}  + \frac{64 H K^2 D^2}{b  (n - 1)}\notag \\
& = \frac{128 e H  K^2 \gamma^{\frac{p}{p+1}}  L_0^{\frac{p}{p+1}} \left(6 R(\wopt)\right)^{\frac{1}{p+1}} }{b  (n - 1)^{1 - p}} + \frac{2 e (64  H K^2 \gamma)^2 L(\wopt) (n-1)^{2p}}{b^2 }  + \frac{2 R(\wopt)}{\gamma (n - 1)^{p+1}}  \notag \\
& ~~~~~ + \frac{2 e (64 H K^2)^2 D^2 \gamma \log\ n}{b^2  (n - 1)^{1 - p}}  + \frac{64 H K^2 \gamma L(\wopt) (n - 1)^{p}}{b}  + \frac{64 H K^2 D^2}{b  (n - 1)}\notag
\intertext{however by condition in Equation \ref{eq:con1}, $\gamma \le \frac{b}{64 H K^2 n^p}$, hence}
& \le \frac{348 H K^2 \gamma^{\frac{p}{p+1}}  L_0^{\frac{p}{p+1}} \left(6 R(\wopt)\right)^{\frac{1}{p+1}} }{b  (n - 1)^{1 - p}}  + \frac{2 e (64 H K^2)^2 D^2 \gamma \log\ n}{b^2  (n - 1)^{1 - p}} \notag\\
& ~~~~~+ \frac{348 H K^2 \gamma L(\wopt) (n - 1)^{p}}{b}  + \frac{2 R(\wopt)}{\gamma (n - 1)^{p+1}}  + \frac{64 H K^2 D^2}{b  (n - 1)} \label{eq:bound}
\end{align}
We shall try to now optimize the above bound w.r.t. $\gamma$, To this end set 
\begin{align}\label{eq:gamma}
\gamma = \min \left\{\frac{1}{4 H},\ \sqrt{\frac{b R(\wopt)}{174 H K^2 L(\wopt) (n-1)^{2p+1}}},\  \left(\frac{b}{1044 H K^2 (n - 1)^{2p}}\right)^{\frac{p+1}{2p+1}} \left(\frac{6 R(\wopt) }{ L_0} \right)^{\frac{p}{2p+1}}\right\}
\end{align}
We first need to verify that this choice of $\gamma$ satisfies the conditions in Equation \ref{eq:con1} and \ref{eq:con2}. To this end, note that as for the condition in Equation \ref{eq:con1}, 
\begin{align*}
\gamma \le \left(\frac{b}{1044 H K^2 (n - 1)^{2p}}\right)^{\frac{p+1}{2p+1}} \left(\frac{6 R(\wopt)}{ L_0} \right)^{\frac{p}{2p+1}} 
\end{align*}
and hence it can be easily verified that for $n \ge 3$, $\gamma \le \frac{b}{64 H K^2 n^p}$. On the other hand to verify the condition in Equation \ref{eq:con2}, we need to show that
\begin{align*}
\gamma & = \min \left\{\frac{1}{4 H},\ \sqrt{\frac{b R(\wopt)}{174 H K^2 L(\wopt) (n-1)^{2p+1}}},\  \left(\frac{b}{1044 H K^2 (n - 1)^{2p}}\right)^{\frac{p+1}{2p+1}} \left(\frac{6 R(\wopt) }{ L_0} \right)^{\frac{p}{2p+1}}\right\}\\
&  \ge \frac{6 R(\wopt)}{n^{p+1} \left(L_0 \right)}
\end{align*}
It can be verified that this condition is satisfied as long as,
\begin{align*}
n \ge \max\left\{3,\ \frac{87 K^2 L(\wopt)}{b} , \frac{783 K^2}{b}  \right\}
\end{align*}
So in effect as long as $n \ge 3$ and sample size $n b \ge \max\{783 K^2, \frac{87 K^2 L(\wopt)}{HD^2}\}$ the conditions are satisfied. Now plugging in this choice of $\gamma$ into the bound in Equation \ref{eq:bound}, we get 
\begin{align*}
a_n &\le  \sqrt{\frac{2784 H K^2 R(\wopt) L(\wopt)}{b (n-1)}} + \frac{2}{3} \left(\frac{6264 H K^2 R(\wopt) L_0^{\frac{p}{p+1}}  }{b (n-1)}\right)^{ \frac{p+1}{2p + 1}}  + \frac{64 H K^2 D^2}{b  (n - 1)} \\
& ~~~~~ + \frac{8 H R(\wopt) }{ (n - 1)^{p+1}} + D^2 \log(n) \left(\frac{64 H K^2  }{b (n - 1)}\right)^{\frac{3p+1}{2p+1}}  \left(\frac{6 R(\wopt) }{ L_0} \right)^{\frac{p}{2p+1}}\\
&\le  \sqrt{\frac{2784 H K^2 R(\wopt) L(\wopt)}{b (n-1)}} + \frac{2}{3} \left(\frac{6264 H K^2 R(\wopt) L_0^{\frac{p}{p+1}}  }{b (n-1)}\right)^{ \frac{p+1}{2p + 1}}  + \frac{64 H K^2 D^2}{b  (n - 1)} \\
& ~~~~~ + \frac{8 H R(\wopt) }{ (n - 1)^{p+1}} + D^2 \log(n) \left(\frac{64 H K^2  }{b (n - 1)}\right)^{\frac{3p+1}{2p+1}}  \left(\frac{6 R(\wopt) }{ L_0} \right)^{\frac{p}{2p+1}}\\
&\le  \sqrt{\frac{2784 H K^2 R(\wopt) L(\wopt)}{b (n-1)}} + \frac{2}{3} \left(\frac{6264 H K^2 R(\wopt) L_0^{\frac{p}{p+1}}  }{b (n-1)}\right)^{ \frac{p+1}{2p + 1}}  + \frac{64 H K^2 D^2}{b  (n - 1)} \\
& ~~~~~ + \frac{8 H R(\wopt) }{ (n - 1)^{p+1}} + \left(\frac{\left(96  K^2\right)^{\frac{p}{p+1}} D^2  }{R(\wopt)}\right)^{\frac{p+1}{2p+1}}  \left(\frac{64 H K^2 R(\wopt) }{b (n - 1)}\right) \frac{\log(n)}{\left(b (n - 1)\right)^{\frac{p}{2p+1}}} \\
& \le \sqrt{\frac{2784 H K^2 R(\wopt) L(\wopt)}{b (n-1)}} + \frac{4176 H K^2  R(\wopt)}{b (n-1)}  \left(\frac{L_0}{6264 H K^2 R(\wopt)}\right)^{\frac{p}{2p+1}}  \left(b (n-1)\right)^{ \frac{p}{2p + 1}} + \frac{64 H K^2 D^2}{b  (n - 1)} \\
& ~~~~~ + \frac{8 H R(\wopt) }{ (n - 1)^{p+1}} +  \left(\frac{64 H K^2 D^2 }{b (n - 1)}\right)  \frac{\log(n)}{\left(b (n - 1)\right)^{\frac{p}{2p+1}} } \left(\frac{384 H K^2  R(\wopt) }{ L_0} \right)^{\frac{p}{2p+1}}\\
& \le \sqrt{\frac{2784 H K^2 R(\wopt) L(\wopt)}{b (n-1)}} + \frac{4176 H K^2  R(\wopt)}{b (n-1)}  \left(\frac{L_0}{6264 H K^2  R(\wopt)}\right)^{\frac{p}{2p+1}}  \left(b (n-1)\right)^{ \frac{p}{2p + 1}} + \frac{64 H K^2 D^2}{b  (n - 1)} \\
& ~~~~~ + \frac{8 H R(\wopt) }{ (n - 1)^{p+1}} +  \left(\frac{64 H K^2 D^2 }{b (n - 1)}\right)  \frac{\log(n)}{\left(b (n - 1)\right)^{\frac{p}{2p+1}} }  
\end{align*}
Picking
$$
p = \min\left\{\max\left\{\frac{\log(b)}{2 \log(n-1)} , \frac{\log \log(n)}{2\left(\log(b(n-1)) - \log \log(n)\right)}\right\} ,1\right\}
$$
we get the bound,
\begin{align*}
a_n &\le \sqrt{\frac{2784 H K^2 R(\wopt) L(\wopt)}{b (n-1)}} + \left(\frac{4176 H K^2  R(\wopt)}{\sqrt{b} (n-1)} + \frac{4176 H K^2  R(\wopt) \sqrt{\log(n)}}{b (n-1)} \right) \left(\frac{L_0}{6264 H K^2 R(\wopt)}\right)^{\frac{1}{3}}  \\
& ~~~~~ + \frac{120 H K^2 D^2}{b  (n - 1)}  + \frac{8 H R(\wopt) }{ (n - 1)^{2}} + \frac{8 H R(\wopt) }{ \sqrt{b} (n - 1)}  +  \frac{64 H K^2 D^2 \sqrt{\log(n)}}{b (n - 1)} \\
&\le \sqrt{\frac{2784 H K^2 R(\wopt) L(\wopt)}{b (n-1)}} + \left(\frac{4176 H K^2  R(\wopt)}{\sqrt{b} (n-1)} + \frac{4176 H K^2  R(\wopt) \sqrt{\log(n)}}{b (n-1)} \right) \left(\frac{L_0}{6264 H K^2 R(\wopt)}\right)^{\frac{1}{3}}  \\
& ~~~~~ + \frac{120 H K^2 D^2}{b  (n - 1)}  + \frac{8 H R(\wopt) }{ (n - 1)^{2}} + \frac{8 H R(\wopt) }{ \sqrt{b} (n - 1)}  +  \frac{64 H K^2 D^2 \sqrt{\log(n)}}{b (n - 1)} \\
&\le \sqrt{\frac{2784 H K^2 R(\wopt) L(\wopt)}{b (n-1)}} + \frac{454  (H K^2 R(\wopt))^{2/3} L_0^{\frac{1}{3}} }{\sqrt{b} (n-1)} + \frac{454 (H K^2 R(\wopt))^{2/3} L_0^{\frac{1}{3}} \sqrt{\log(n)}}{b (n-1)}   \\
& ~~~~~ + \frac{120 H K^2 D^2}{b  (n - 1)}  + \frac{8 H R(\wopt) }{ (n - 1)^{2}} + \frac{8 H R(\wopt) }{ \sqrt{b} (n - 1)}  +  \frac{64 H K^2 D^2 \sqrt{\log(n)}}{b (n - 1)} 
\end{align*}
Recall that $L_0 = \frac{3}{2} H D^2 + L(\wopt)$. Now note that if $L(\wopt) \le H K^2 D^2/2$ then $L_0 \le 2 H K^2 D^2$, on the other hand if $L(\wopt) > H K^2 D^2/2$ then $(H K^2 R(\wopt))^{2/3} L_0^{\frac{1}{3}} \le \sqrt{4 H K^2 R(\wopt) L(\wopt)}$. Hence we can conclude that,
\begin{align*}
a_n &\le \sqrt{\frac{2784 H K^2 R(\wopt) L(\wopt)}{b (n-1)}} + \frac{454  H K^2  (R(\wopt))^{2/3} (2 D^2)^{\frac{1}{3}} }{\sqrt{b} (n-1)} + \frac{454 H K^2 (R(\wopt))^{2/3} (2 D^2)^{\frac{1}{3}} \sqrt{\log(n)}}{b (n-1)}   \\
& ~~~~~ + \frac{120 H K^2 D^2}{b  (n - 1)}  + \frac{8 H R(\wopt) }{ (n - 1)^{2}} + \frac{8 H R(\wopt) }{ \sqrt{b} (n - 1)}  +  \frac{64 H K^2 D^2 \sqrt{\log(n)}}{b (n - 1)} + \frac{908 \sqrt{H K^2 R(\wopt) L(\wopt)} }{\sqrt{b} (n-1)} \\
& ~~~~~ + \frac{908 \sqrt{H K^2 R(\wopt) L(\wopt) \log(n)}}{b (n-1)} 
\end{align*}
Since $n > 783 K^2$ and $R(\wopt) \le D^2/2$ we can conclude that
\begin{align*}
a_n &\le 164 \sqrt{\frac{H K^2 R(\wopt) L(\wopt)}{b (n-1)}} + \frac{580  H K^2  (R(\wopt))^{2/3}  D^{\frac{2}{3}} }{\sqrt{b} (n-1)}   +  \frac{545 H K^2 D^2 \sqrt{\log(n)}}{b (n - 1)}  + \frac{8 H R(\wopt) }{ (n - 1)^{2}} 
\end{align*}
This concludes the proof.
\end{proof}

\begin{proof}[Proof of Theorem \ref{thm:ag}]
For Euclidean case $R(\w) = \frac{1}{2} \norm{\w}_2^2$ and $K = \sqrt{\sup_{\w : \norm{\w}_2 \le 1} \norm{\w}^2} = 1$. 
Plugging these in the previous theorem (along with appropriate step size) we get
\begin{align*}
\E{L(\w^{\mathrm{ag}}_n)} - L(\wopt) &\le 116 \sqrt{\frac{H \norm{\wopt}^2 L(\wopt)}{b (n-1)}} + \frac{366  H  \norm{\wopt}^{4/3}  D^{\frac{2}{3}} }{\sqrt{b} (n-1)}   +  \frac{545 H D^2 \sqrt{\log(n)}}{b (n - 1)}  + \frac{4 H \norm{\wopt}^2 }{ (n - 1)^{2}} 
\end{align*}
The second inequality is a direct consequence of the fact that $\norm{\wopt} \le D$.
\end{proof}

\subsection{Some Technical Lemmas}

\begin{lemma}\label{cor:eqsmooth}
Denote $K:= \sqrt{2 \sup_{\w : \norm{\w} \le 1} R(\w)}$, then for any $\x_{1},\ldots,\x_b$ mean zero vectors drawn iid from any fixed distribution, 
$$
\E{\norm{\frac{1}{b} \sum_{t=1}^b \x_t}_*^2} \le \frac{K^2}{b^2} \sum_{t=1}^b \E{\|\x_t\|_*^2}
$$
\end{lemma}
\begin{proof}
We start by noting that
\begin{align}
\norm{\frac{1}{b} \sum_{t=1}^i \x_t}_*^2 & =  \left( \sup_{\w :\|\w\| \le 1 }\ip{\w}{\frac{1}{b} \sum_{t=1}^i \x_t } \right)^2 \notag \\
 & =  \left(\inf_{\alpha} \frac{1}{\alpha} \sup_{\w :\|\w\| \le 1 }\ip{\w}{ \frac{\alpha}{b} \sum_{t=1}^i \x_t } \right)^2 \notag \\
 & \le \left(\inf_{\alpha}\left\{  \frac{1}{\alpha} \sup_{\w :\|\w\| \le 1 } R(\w) + \frac{1}{\alpha} R^*\left(\frac{\alpha}{b} \sum_{t=1}^i \x_t\right) \right\}\right)^2 \notag \\ 
  & = \left(\inf_{\alpha}\left\{  \frac{K^2}{2 \alpha}  + \frac{1}{\alpha} R^*\left(\frac{\alpha}{b} \sum_{t=1}^i \x_t\right) \right\}\right)^2 \label{eq:nrmsmt}
\end{align}
where the step before last was due to Fenchel-Young inequality and $R^*$ is simply the convex conjugate of $R$. Now For any $i \in [b]$ define $S_i = R^*\left(\frac{\alpha}{b} \sum_{t=1}^i \x_t\right)$. We claim that 
\begin{align*}
\E{S_i} \le \E{S_{i-1}} + \frac{\alpha^2}{2 b^2}\E{\|\x_i\|_*^2}
\end{align*}
To see this note that since $R$ is $1$-strongly convex w.r.t. $\norm{\cdot}$, by duality $R^*$ is $1$-strongly smooth w.r.t. $\norm{\cdot}_*$ and so for any $i \in [b]$,
\begin{align*}
R^*\left(\frac{1}{b} \sum_{t=1}^i \x_t\right) & \le R^*\left(\frac{1}{b} \sum_{t=1}^{i-1} \x_t\right) + \frac{1}{2 b} \ip{\nabla R^*\left(\frac{1}{b} \sum_{t=1}^{i-1} \x_t\right)}{\x_i} + \frac{\alpha^2}{2 b^2} \norm{\x_i}_*^2
\end{align*}
taking expectation w.r.t. $\x_i$ and noting that $\E{\x_i} = 0$ by assumption we see that
\begin{align*}
\Es{\x_b}{S_i} & \le S_{i-1}  + \frac{\alpha^2}{2 b^2} \Es{\x_i}{\norm{\x_i}_*^2}
\end{align*}
Taking expectation we get as claimed that : 
$$
\E{S_i} \le \E{S_{i-1}} + \frac{\alpha^2}{2 b^2}\E{\|\x_i\|_*^2}
$$
Now using this above recursively (and noting that $S_0 = 0$ ) we conclude that 
$$
\E{S_i} \le \frac{\alpha^2}{2 b^2} \sum_{t=1}^{i} \E{\|\x_t\|_*^2}
$$
Plugging this back in Equation \ref{eq:nrmsmt} we get
\begin{align*}
\E{\norm{\frac{1}{b} \sum_{t=1}^b \x_t}_*^2} & \le \left(\inf_{\alpha}\left\{  \frac{K^2}{2 \alpha}  + \frac{\alpha}{2 b^2} \sum_{t=1}^{i} \E{\|\x_t\|_*^2} \right\}\right)^2\\
& = \left(\inf_{\alpha}\left\{  \frac{K^2}{2 \alpha}  + \frac{\alpha}{2 b^2} \sum_{t=1}^{i} \E{\|\x_i\|_*^2} \right\}\right)^2 = \frac{K^2}{b^2} \sum_{t=1}^{i} \E{\|\x_t\|_*^2}
\end{align*}
\end{proof}

\begin{lemma}\label{ut:rec}
Consider a sequence of non-negative number $a_{1},\ldots,a_n \in [0,a_0]$ that satisfy
$$
a_n \le A(n) \sum_{i=1}^{n-1} a_i + B(n)
$$
where $A$ is decreasing in $n$. For such a sequence, for any $m \in [n]$, as long as $A(i) \le 1/2$ for any $i \ge n-m-1$ and  $\sum_{i=n-m-1}^{n} A(i)  \le 1$ then
\begin{align*}
a_n  & \le e A(n)\left( a_0 (n-m)  +  \sum_{i= n-m-1}^n B(i) \right) + B(n) 
\end{align*}
\end{lemma}
\begin{proof}
We shall unroll this recursion. Note that
\begin{align*}
a_n & \le A(n) \sum_{i=1}^{n-1} a_i + B(n)\\
& = A(n) \left( \sum_{i=1}^{n-2} a_i +  a_{n-1} \right) + B(n)\\
& \le  A(n) \left( \sum_{i=1}^{n-2} a_i + A(n-1) \sum_{i=1}^{n-2} a_i + B(n-1) \right) + B(n)\\ 
& = A(n) (1 + A(n-1)) \sum_{i=1}^{n-2} a_i + B(n) + A(n) B(n-1)\\
& \le A(n) (1 + A(n-1)) \left(\sum_{i=1}^{n-3} a_i + A(n-2) \sum_{i=1}^{n-3} a_i + B(n-2) \right) + + B(n) + A(n) B(n-1)\\
& = A(n) (1 + A(n-1)) (1 + A(n-2)) \sum_{i=1}^{n-3} a_i + B(n) + A(n) B(n-1) + A(n) (1 + A(n-1)) B(n-2)
\end{align*}
Continuing so upto $m$ steps we get
\begin{align}\label{eq:rec}
a_n \le A(n) \left( \prod_{i=1}^{m-1} (1 + A(n-i)) \right) \sum_{i=1}^{n-m} a_i + B(n) + A(n) \left( \sum_{i=1}^{m-1} \left( \prod_{j=1}^{i-1} (1 + A(n-j)) \right) B(n-i) \right)
\end{align}

We would now like to bound in general  the term $\prod_{i=1}^{m-1} (1 + A(n-i)) $. To this extant note that,
$$
\prod_{i=1}^{m-1} (1 + A(n-i))  = \exp\left( \sum_{i=1}^{m-1} \log(1 + A(n-i))\right)
$$
Now assume $A(i) \le 1/2$ for all $i \ge n-m-1$ so that $\log(1 + A(n-i)) \le A(n-i)$. We get 
$$
\prod_{i=1}^{m-1} (1 + A(n-i))  \le \exp\left( \sum_{i=1}^{m-1} A(n-i)\right)
$$
Now if $\sum_{i=n-m-1}^{n} A(i)  \le 1$ then we can conclude that
$$
\prod_{i=1}^{m-1} (1 + A(n-i))  \le e
$$
Plugging this in Equation \ref{ut:rec} we get
\begin{align*}
a_n & \le e A(n)\left( \sum_{i=1}^{n-m} a_i   + \sum_{i=1}^{m-1} B(n-i) \right) + B(n) \\
& = e A(n)\left( \sum_{i=1}^{n-m} a_i   + \sum_{i= n-m-1}^n B(i) \right) + B(n) 
\end{align*}

Now if for each $i \le n$, $a_i \le a_0$ then we see that
\begin{align*}
a_n  & \le e A(n)\left( a_0 (n-m)  + \sum_{i= n-m-1}^n B(i)  \right) + B(n) 
\end{align*}
Hence we conclude that as long as $\sum_{i=n-m-1}^{n} A(i)  \le 1$
\begin{align*}
a_n  & \le e A(n)\left( a_0 (n-m) + \sum_{i= n-m-1}^n B(i) \right) + B(n) 
\end{align*}
\end{proof}

\end{document}